\documentclass[10pt,twocolumn,letterpaper]{article}

\usepackage[pagenumbers]{cvpr}              

\usepackage{comment}
\usepackage{gensymb}
\usepackage{multirow}
\usepackage{subcaption}
\usepackage{adjustbox}
\usepackage{enumitem}
\newcommand{\name}{EgoExoMoCap}

\definecolor{blue}{HTML}{0071BC}
\definecolor{yellow}{HTML}{EA9D00}
\definecolor{gray}{HTML}{4d4d4d}

\newcommand{\np}[1]{%
  \vspace{1.0ex}%
  \noindent\textit{#1}\enspace\ignorespaces
}

\usepackage{xifthen}

\newcommand{\normal}[3][]{%
    \ifthenelse{\isempty{#1}}
        {\mathcal{N}\left(#2, #3\right)}
        {\mathcal{N}\left(#1|#2, #3\right)}%
}

\newcommand{\vect}[1]{\mathbf{#1}}

\makeatletter
\renewcommand{\abstract}[1]{\centerline{\large\bf Abstract}\vspace*{12pt}\noindent{\it #1}\vspace*{12pt}}
\makeatother

\definecolor{cvprblue}{rgb}{0.21,0.49,0.74}
\usepackage[pagebackref,breaklinks,colorlinks,citecolor=cvprblue]{hyperref}

\def\paperID{0000}
\def\confName{CVPR}
\def\confYear{2026}

\title{EgoExoMoCap: Distributed Ego-Exo Human Motion Capture}

\author{%
Jiaxi Jiang\textsuperscript{1,2*}\quad
Bharat Lal Bhatnagar\textsuperscript{1}\quad
Nan Yang\textsuperscript{1}\quad
Lingni Ma\textsuperscript{1}\quad
Sebastian Starke\textsuperscript{1}\\
Robin Kips\textsuperscript{1}\quad
Nadine Bertsch\textsuperscript{1}\quad
Christian Holz\textsuperscript{2}\quad
Federica Bogo\textsuperscript{1}\\[3pt]
\textsuperscript{1}Meta Reality Labs \quad \textsuperscript{2}ETH Zürich\\[2pt]
\url{https://siplab.org/projects/EgoExoMoCap}
}

\newcommand{\authnote}[1]{\begingroup\renewcommand{\thefootnote}{*}\footnotetext{#1}\endgroup}
\begin{document}
\twocolumn[{%
  \renewcommand\twocolumn[1][]{#1}%
    \maketitle
      \begin{center}
          \includegraphics[width=\linewidth]{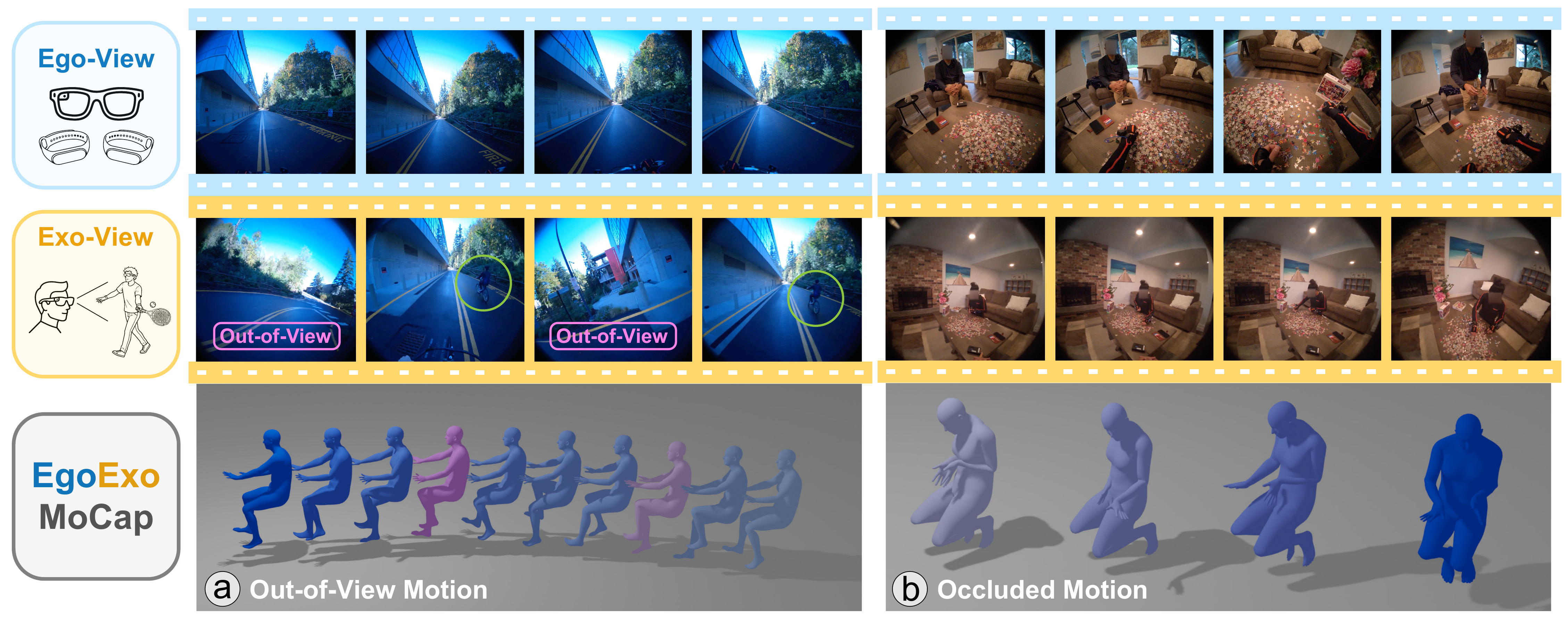}
              \captionof{figure}{\textcolor{blue}{Ego}\textcolor{yellow}{Exo}\textcolor{gray}{MoCap} is a lightweight, distributed human motion capture system. Given two or more subjects wearing head-mounted devices, the approach combines continuous \textcolor{blue}{egocentric} signals with intermittent \textcolor{yellow}{exocentric} camera views to robustly handle challenges such as out-of-view motions (a) and severe body occlusions (b).}
                  \label{fig:teaser}
                    \end{center}%
                      \vspace{4pt}%
                      }]
\authnote{This work was done during an internship at Meta.}

        \abstract{
Human motion capture from head-mounted devices (HMDs) offers a scalable way to acquire real-world human motion and interaction data, 
which is crucial for applications in embodied AI and VR/AR.
Existing approaches focus on either egocentric body tracking, estimating the motion of the subject wearing the device, or exocentric tracking, capturing the movements of people in the wearer’s surroundings. So far, these two paradigms have largely been explored in isolation.
In this paper, we propose a novel distributed framework that jointly leverages ego- and exocentric multi-modal signals for human motion estimation from HMDs.
Unlike traditional motion capture systems requiring bulky multi-camera setups or obtrusive mocap suits, our approach, \name{}, is as simple as two (or more) people, each wearing a pair of smart glasses. 
The method leverages head (plus potentially wrist) tracking signals for accurate estimation of global motion in the 3D world and combines context-aware image features based on DINOv3 to achieve robustness in the presence of noise and occlusions.
Extensive experiments on two in-the-wild datasets show that our approach can robustly reconstruct motion even in challenging scenarios. 
}
        \section{Introduction}
\label{sec:intro}
Accurate and robust human motion capture in the wild is important for many applications, including robotics, virtual and augmented reality (VR/AR), and human-computer interaction.
Recent efforts have shown growing interest in using egocentric devices and videos as scalable sources of human demonstrations~\cite{punamiya2026egoverse,yang2025egovlalearningvisionlanguageactionmodels,kareer2025egomimic,hoque2025egodex,shi2026egohumanoid}, yet accurate full-body motion capture from such devices remains challenging, as traditional motion capture systems often require bulky multi-camera setups or obtrusive mocap suits.

Several approaches in the literature~\cite{shin2024wham,wang2024tram,shen2024gvhmr,wang2025prompthmr} focus on in-the-wild capture via \emph{exocentric} body tracking: an external RGB camera (\eg, from a phone) captures the performance of the subject and the resulting monocular video is used to infer their 3D motion.
These methods often struggle in reconstructing global motion in world space, given the inherent ambiguity of the problem; furthermore, they suffer in the presence of body occlusions, fast camera motions, and image blur.

An alternative is offered by the recent proliferation of wearable devices, such as smart glasses equipped with cameras and inertial sensors (\eg, Project Aria~\cite{engel2023project}). Lightweight and easily usable for hours, these devices capture multi-modal streams including head and hand trajectories plus stereo/RGB images.
%
Recent work~\cite{jiang2022avatarposer,jiang2024egoposer,guzov2024hmd,barquero2025sparse} leverage these devices for \emph{egocentric} body motion tracking and synthesis. Approaches commonly rely on head and wrist poses, plus potentially egocentric camera streams, to infer the wearer's pose. However, sensor noise and the limited visibility of the subject's body from the egocentric cameras make it difficult to faithfully reconstruct motions.

Recent progress in multi-human motion estimation~\cite{xue2025group,multihmr,multiphys,chi2024m2d2m} highlights the need to capture coordinated human behaviors and interactions. 
However, existing methods mainly estimate people from external observations or individually worn sensors, without exploiting mutual observations among multiple users. More broadly, exocentric and egocentric sensing has been largely studied in isolation, with only a few efforts using third-person or external views as training-time supervision for egocentric pose estimation~\cite{dhamanaskar2023enhancing,wang2022estimating}. 
In contrast, a multi-HMD setup turns each participant into both a motion subject and a mobile observer of others, creating a natural opportunity for collaborative motion capture. 
Enabled by modern HMDs that support distributed information exchange and accurate time alignment~\cite{aria2}, we introduce \name{}, a unified framework that jointly leverages egocentric self-motion cues and exocentric tracking signals from Aria glasses~\cite{engel2023project} for distributed human motion capture.

The first challenge is how to effectively combine heterogeneous signals: the head (plus potentially wrist) poses tracked by the glasses worn by one subject (\emph{wearer}), plus the images coming from the glasses worn by another subject \emph{observing} the wearer. Our approach first estimates an initial body pose from egocentric streams, which serves as initialization and provides reliable region proposals for person localization in exocentric views. Subsequently, we detect 2D keypoints~\cite{vitpose} in the exocentric view, unproject them to 3D rays via the observer's camera parameters, and transform them into the egocentric coordinate frame via the wearer's camera parameters. We find this a simple yet effective solution to unify input streams into a consistent (egocentric) coordinate frame. It generalizes well across diverse motions and ensures scalability to scenarios encompassing multiple observers/exocentric images.

The second challenge is posed by the unreliability of exocentric streams. As the observer looks around, the wearer might be partially or totally out of the field of view. Frequent human-scene interactions, also studied in prior work on scene-aware motion modeling and human-scene interaction~\cite{zhang2023probabilistic,araujo2023circle,li2023object,shi2025caring}, cause additional body occlusions and pose ambiguities; in these scenarios, even state-of-the-art 2D keypoint estimators are not robust enough. We propose to leverage the context around the subject, computing deep image features with DINOv3~\cite{simeoni2025dinov3} and using them to learn per-keypoint confidence scores.
This effectively balances the contributions of egocentric and exocentric signals, relying more on one or the other depending on the reliability of the exocentric streams. 

To summarize, our contributions are as follows:

(1) We propose a lightweight, portable solution for in-the-wild motion capture, which just relies on a set of people wearing HMDs such as Aria glasses~\cite{engel2023project}.

(2) We design a multi-modal framework that combines heterogeneous ego- and exocentric signals, such as continuous head trajectories and intermittent image features, to ensure robust motion estimates. The approach works with as few as two subjects and naturally scales to multi-subject setups.

(3) We evaluate our approach on indoor and outdoor sequences from two in-the-wild datasets, Nymeria~\cite{ma2024nymeria} and EgoHumans~\cite{egohumans}, achieving state-of-the-art performance in real-world scenarios.

By reducing reliance on centralized multi-camera setups and obtrusive motion-capture suits, EgoExoMoCap makes full-body motion capture more accessible outside controlled studios, enabling scalable collection of real-world whole-body motion and interaction data for embodied AI, VR/AR, and interactive agents.

        \section{Related Work}
\label{sec:related_work}

\noindent\textbf{Egocentric human motion estimation.}
The problem of full-body pose reconstruction from HMDs has received growing attention in the past years~\cite{grauman2024ego,ma2024nymeria,zhang2022egobody,jiang2022avatarposer,jiang2024egoposer,liu20214d}. 
AvatarPoser~\cite{jiang2022avatarposer} introduced a Transformer-based framework for full-body pose estimation from sparse HMD tracking signals, demonstrating that plausible articulated body motion can be recovered from only head and wrist observations. Follow-up work~\cite{zheng2023realistic} improves robustness via a two-stage framework leveraging body joint correlations. 
QuestSim~\cite{winkler2022questsim,lee2023questenvsim} and SimXR~\cite{Luo_2024_CVPR} use physics simulation to generate plausible motions. 
MANIKIN~\cite{jiang2024manikin} combined neural networks with analytical inverse kinematics, using biomechanical constraints and predicted swivel angles~\cite{tolani2000real} to recover full-body poses
There have also been a number of generative approaches based on VAEs~\cite{dittadi2021full}, normalizing flows~\cite{aliakbarian2022flag}, VQ-VAEs~\cite{starke2024categorical,egolm,feng2023sage}, and diffusion~\cite{du2023avatars,dong2024,castillo2023bodiffusion}.
Most of these approaches assume head and wrist trajectories are always available as input.
However, lightweight wearable glasses, when not accompanied by additional sensors like wristbands~\cite{ma2024nymeria}, cannot provide reliable wrist trajectories~\cite{jiang2024egoposer,aliakbarian2023nemo,chi2024estimating}. EgoPoser~\cite{jiang2024egoposer} proposes a system that is robust to intermittent hand tracking signals and can work in large scenes via global motion decomposition, while 
predicting also body shape. 
Similarly, DSPoser~\cite{chi2024estimating} and EgoAllo~\cite{yi2025estimating} use off-the-shelf hand pose estimators to predict hand motions, which then guide full-body motion estimation.
EgoEgo~\cite{li2023ego} uses head motion, while HMD2~\cite{guzov2024hmd} relies on head trajectories plus egocentric camera streams.
RPM~\cite{barquero2025sparse} proposes a temporally causal rolling prediction framework that produces smoother hand trajectories.
All these approaches deal with the limitation of just using egocentric signals: while they can synthesize plausible motions, they can hardly reconstruct lower body movement in a faithful way.

\noindent\textbf{Exocentric human motion estimation.}
There is a rich literature on human pose and shape (HPS) estimation from images~\cite{hmr,spin,refit,pare,hmr2,pymaf,cliff,I2l,pose2mesh,SPEC:ICCV:2021,meshgraphform}. In the following, we focus in particular on monocular human motion estimation from unconstrained, monocular videos.
Several approaches~\cite{vibe,dsd,hmmr,Luo_2020_ACCV,tcmr} propose to combine 3D human pose estimates (either 3D joints or body model parameters~\cite{smplx}) with temporal models to reconstruct smooth motions. Since camera extrinsics over time are in general unknown, these methods focus on retrieving 3D body motion in camera space -- without returning coherent global motion in world coordinates.
Recent methods try to overcome this limitation, considering dynamic camera captures and typically following a two-stage approach: they first estimate camera parameters via SLAM~\cite{droid,bodyslam,bodyslam++,teed2024dpvo}, and then leverage human motion priors to optimize pose in world coordinates~\cite{pace,slahmr,glamr}. Other approaches~\cite{shin2024wham,shen2024gvhmr,genmo2025} train temporal models to directly regress global human motion from image and camera features. Others~\cite{wang2024tram,Zhao2024globalcamhuman} solve for scale ambiguities via monocular metric depth.
Most approaches assume the human body is fully visible in most frames -- which often does not hold in real-world scenarios~\cite{harmony4d}. Some recent work tries to directly handle body occlusions~\cite{zhang2024rohm}, also considering HMD exocentric images~\cite{zhang2022egobody}. LAMP~\cite{yang2026lamp}
further leverages localized multi-camera HMD input for metric 3D people tracking in world coordinates.
We propose to overcome the challenges of exocentric motion estimation by effectively leveraging the multi-modal streams offered nowadays by HMDs.

\noindent\textbf{Human motion capture with body-worn sensors.}
There is a rich literature on reconstructing body motions from body-worn inertial sensors~\cite{dynaip2024,Van_Wouwe_2024_CVPR,Zuo_2024_CVPR_loose,yi2024pnp,yi2025improving,von2017sparse,huang2018deep,ilic2025human}.
A well-known challenge in these pipelines is drift, caused by the absence of absolute positioning data. Approaches in the literature try to tackle this challenge by combining sensor data with additional multi-modal streams~\cite{chen2024motion}.
RGB cameras are typically used to obtain more robust results: \cite{von2018recovering,pan2023fusingmono} combine body-worn IMUs with body images captured with a moving camera; \cite{daip2024hmdposer,liu2024egohdm} leverage IMUs and an egocentric camera, also achieving 3D scene reconstruction. 
EgoSim~\cite{hollidt2024egosim} simulates multiple body-worn cameras.
Other approaches also leverage depth and plantar pressure sensors~\cite{zhang2024mmvp}, LiDAR and event cameras~\cite{yan2024reli11d}, and Ultra-Wideband Units~\cite{UIP,liu2025umotion,xue2025group}.
An interesting line of work focuses on motion capture ``anywhere'', by combining multi-modal streams from consumer devices~\cite{wang2026embodmocap}. EgoFormer~\cite{egohumans} tracks humans based on RGB and grayscale images captured with Aria glasses. 
IMUPoser~\cite{mollyn2023imuposer} leverages IMUs from smartphones, smartwatches, and earbuds. \cite{Lee_2024_CVPR} relies on two smartwatches and a head-mounted camera.
Similar in spirit to this line of work, our approach proposes a lightweight, easily portable system; in particular, our framework is distributed and scalable -- exploiting multiple devices worn by two or more people. 

        \begin{figure*}[t]
    \centering
    \includegraphics[width=1\linewidth]{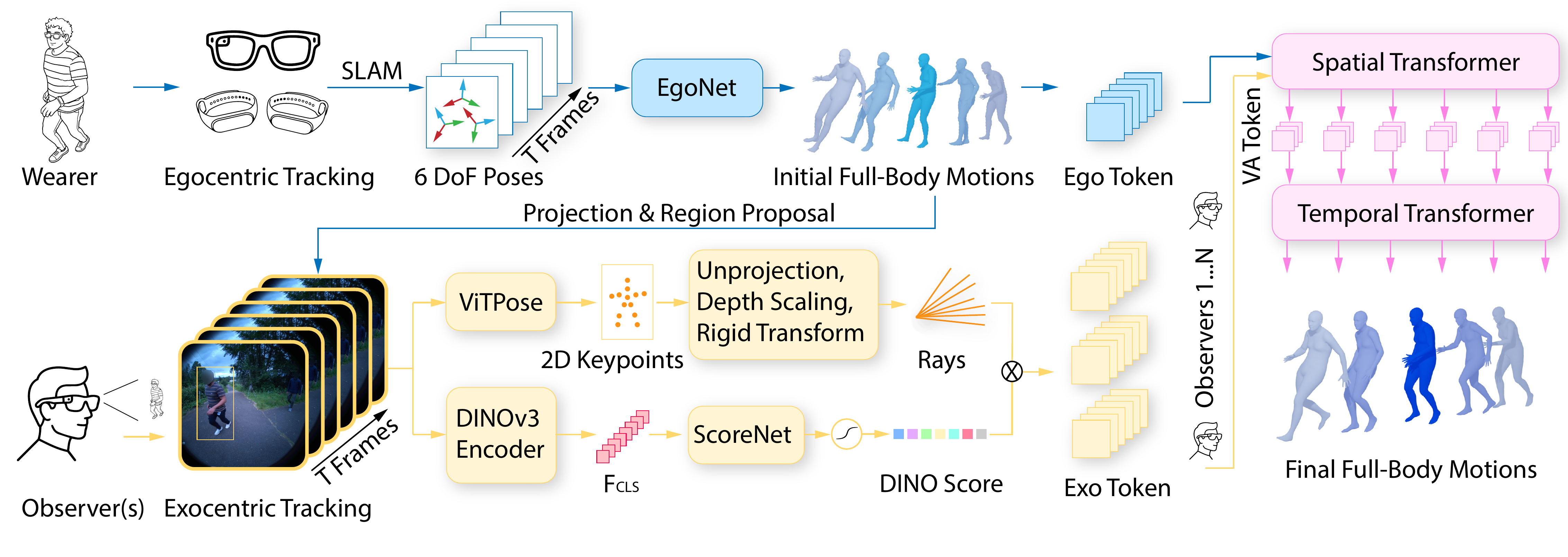}
    \caption{\textbf{Overview of \name{}.} Given an egocentric and one or more exocentric streams from HMDs, we first roughly estimate 3D body poses from egocentric streams (EgoNet) to identify regions of interest in exocentric frames.
    From these, ViTPose~\cite{vitpose}-extracted 2D keypoints are unprojected into 3D rays and softly weighted by DINOv3~\cite{simeoni2025dinov3}-based confidence scores to form Exo Tokens. A Spatial Transformer fuses Ego and Exo tokens into View-Aggregated (VA) Tokens, followed by a Temporal Transformer for smoothness to output final full-body motions.
    }
    \label{fig:pipeline}
\end{figure*}
\section{Method}
\label{sec:method}

\subsection{Problem Formulation}
We focus on full-body motion capture in a distributed setup, where two or more subjects wear an HMD equipped with inertial sensors and exocentric cameras (\eg, Aria glasses~\cite{engel2023project}). For simplicity, in the following we describe a two-person scenario, involving a \emph{wearer} (target subject whose motion needs to be reconstructed) and an \emph{observer} (nearby subject looking at the wearer). Our approach can be extended to scenarios with more observers.

\np{Inputs.} At each timestep $t \in \{1, \dots, T\}$, our method takes as input:
\begin{itemize}
\item an RGB image $\vect{I}_t^o \in \mathbb{R}^{H \times W \times 3}$ from the observer's head-mounted camera,
    \item the observer's head position $\vect{p}_t^{o} \in \mathbb{R}^3$ and orientation $\boldsymbol{\theta}_t^{o} \in \mathbb{R}^6$ (in 6D representation~\cite{zhou2019continuity}),
    \item the wearer's head position $\vect{p}_t^{w,\text{head}} \in \mathbb{R}^3$ and orientation $\boldsymbol{\theta}_t^{w,\text{head}} \in \mathbb{R}^6$,
    \item optionally, position $\vect{p}_t^{w,\text{lrw}}$ and orientation $\boldsymbol{\theta}_t^{w,\text{lrw}}$ of the wearer's left and right wrists. 
\end{itemize}
Head trajectories (positions and orientations) can be obtained from visual-inertial SLAM systems built into HMDs~\cite{engel2023project,apple_vision_pro,hololens}. Wrist trajectories can be obtained, for example, either from camera-equipped wristbands using SLAM~\cite{ma2024nymeria} or from handheld controllers commonly used in VR systems~\cite{quest}.
We consider both the case in which wrist trajectories are available (\textbf{3-point}) and the one in which only the head trajectory is known (\textbf{1-point}).
We assume camera extrinsics and intrinsics parameters are known for both the wearer and the observer (HMDs usually provide factory calibration information).

\np{Outputs.} Our model predicts the sequence of full-body poses of the wearer in global space. We use the SMPL body model~\cite{loper2015smpl}, parameterized by pose parameters $\boldsymbol{\theta}_t^{w,\text{body}} \in \mathbb{R}^{J \times 6}$ (local joint rotations of $J{=}21$ joints, in 6D representation), together with root joint global orientation $\boldsymbol{\theta}_t^{w,\text{root}} \in \mathbb{R}^6$ and position $\vect{p}_t^{w,\text{root}} \in \mathbb{R}^3$. 
Note that the model does not predict SMPL shape parameters, but robustly handles different shapes as input -- either subject-specific~\cite{barquero2025sparse} or corresponding to the SMPL mean identity~\cite{yi2025estimating}. We obtain
3D joint positions at each timestamp $t$ via forward kinematics.

\begin{figure*}[t]
    \centering
    \includegraphics[width=0.95\linewidth]{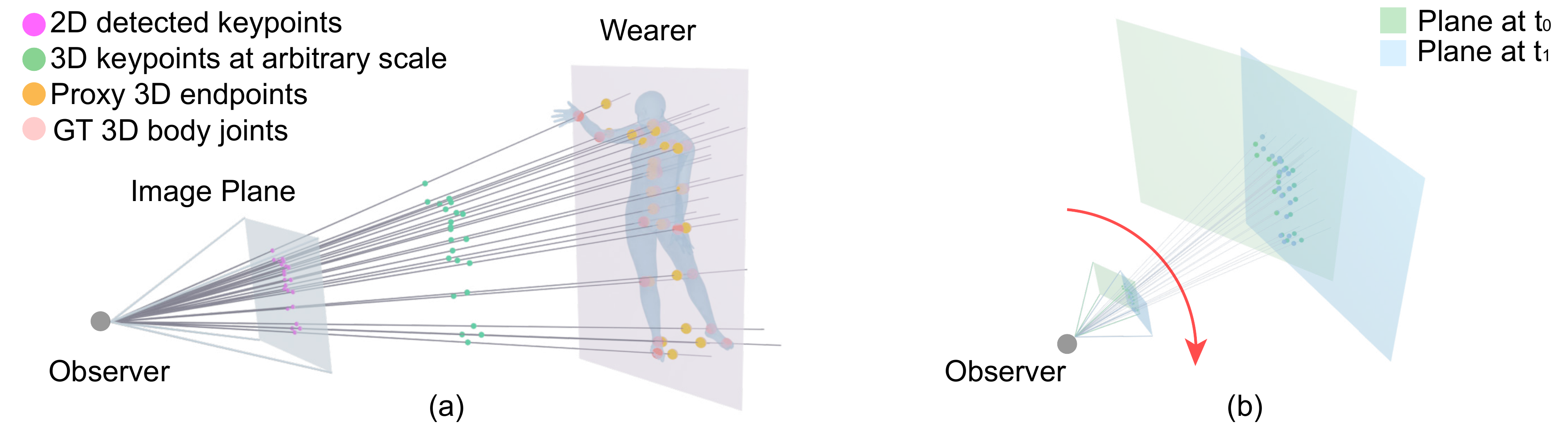}
    \caption{\textbf{Ray-based representation.} (a) We unproject 2D keypoints into 3D rays that can intersect planes at different depths (green, yellow).
    By using the wearer-observer distance, we obtain the correct depth and correctly scaled 3D endpoints (yellow). (b) Leveraging the observer's camera extrinsics ensures head-rotation invariance: proxy endpoints are the same if the observer moves but the wearer is still. }
    \label{fig:ray}
\end{figure*}

\subsection{Method Overview}

\Cref{fig:pipeline} provides an overview of our framework. Given the wearer's egocentric input streams, we employ a temporal network (EgoNet) to coarsely estimate full-body poses and project them onto the observer's exocentric images to localize the wearer and produce region proposals (\Cref{sec:wearer_localization}). Within this region, we extract 2D body keypoints~\cite{vitpose}. To account for different exo views (multiple observers) and observers' head motion, we lift 2D keypoints to 3D rays and transform them into the wearer's egocentric coordinate frame (\Cref{sec:ray_canonicalization}). Since 2D keypoints (and therefore rays) are unreliable under occlusion, we leverage DINOv3 features~\cite{simeoni2025dinov3} to associate keypoint predictions with learned confidence scores (\Cref{sec:visibility_gating}). The resulting \emph{exo} tokens are combined with \emph{ego} ones (\ie, egocentric input streams and EgoNet output) and processed with spatial and temporal transformers to predict the final body poses (\Cref{sec:fusion}).

\subsection{Ego-Guided Wearer Localization}
\label{sec:wearer_localization}

\np{Egocentric signal encoding.}
At each timestep $t$, we encode the wearer's head position $\vect{p}_t^{w,\text{head}}$ and orientation $\boldsymbol{\theta}_t^{w,\text{head}}$, along with their velocities, into a 18D vector. When wrist tracking is available, we additionally incorporate wrist positions $\vect{p}_t^{w,\text{lrw}}$ and orientations $\boldsymbol{\theta}_t^{w,\text{lrw}}$, together with their velocities, extending the representation to 54D.
We apply spatial normalization as in~\cite{jiang2022avatarposer} and express wrist coordinates relative to the head, removing global position dependence; we also temporally normalize the head position at each frame $t > 1$ relative to the first frame ($t{=}1$), factoring out absolute starting location. 
After normalization, we include a 6D relative displacement encoding -- expressing the head displacement on the XY plane at time $t$ with respect to the position at timestep $1$ -- yielding a 60D egocentric feature vector $\vect{x}_t^{\text{ego}} \in \mathbb{R}^{60}$.

\np{Coarse pose estimation and region proposal.} To extract pose cues from exocentric images, we need to roughly localize the wearer in them. We observe that common bounding box detectors~\cite{vitpose,yolo} do not work well in scenarios with strong occlusions and viewpoint changes.
Therefore we train an ego-only temporal network \textbf{EgoNet} that takes $\vect{x}_t^{\text{ego}}$ as input and predicts initial SMPL pose parameters (both local and global). 
EgoNet linearly projects the egocentric signals to a 512D hidden state with a sinusoidal temporal positional encoding~\cite{vaswani2017attention}, applies a single MLP-Mixer block~\cite{tolstikhin2021mlp} consisting of a token-mixing MLP across the time dimension for temporal dependencies followed by a channel-mixing MLP across features, and uses two MLP heads to predict global orientation and local body poses.

While coarse, these estimates provide a reasonable initialization.
From them, we compute 3D joint positions in global space and project them on the exocentric image via the observer's camera parameters. 
The bounding box computed from the projected joints, expanded by a fixed margin, defines a region of interest. Subsequent exocentric processing -- 2D keypoints (\Cref{sec:ray_canonicalization}) and DINOv3 features (\Cref{sec:visibility_gating}) -- 
operates within this region, ensuring that the wearer is consistently tracked across frames.

\subsection{Exocentric Ray-based Pose Canonicalization}
\label{sec:ray_canonicalization}

Leveraging the bounding boxes obtained via EgoNet poses, we run ViTPose~\cite{vitpose} to detect $K=13$ body 2D keypoints, 
representing the wearer's 2D pose in the observer's image. Directly using these 2D coordinates as network input would inherently entangle the observer's viewpoint and camera parameters, limiting cross-scenario generalization. 
To reduce this camera dependence, we adopt a ray-based geometric representation,
following camera-aware 3D pose formulations~\cite{zhan2022ray3d,cho2021camera}.
Inspired by LAMP~\cite{yang2026lamp}, which lifts 2D keypoints into
3D exocentric ray clouds using known HMD poses and calibration, we further extend this representation and develop a wearer-conditioned ray formulation for ego-exo fusion: exocentric keypoints are lifted with the observer pose, scaled by the observer--wearer head distance, and
canonicalized in the wearer's head-local frame before being fused with continuous
egocentric tracking signals,
as illustrated in Fig.~\ref{fig:ray}.

\np{From 2D detections to 3D rays.} 
Given a detected 2D keypoint $\vect{x}_{t,j}$ at time $t$, we utilize the observer's camera intrinsics $\vect{K}_{obs}$ and extrinsics rotation $\vect{R}_{obs}$ 
to unproject it into a 3D ray in global coordinates, then normalized into a unit vector (representing a direction):
\begin{equation}
    \hat{\vect{d}}_{t,j} = \frac{\vect{R}_{obs} \vect{K}_{obs}^{-1} \vect{x}_{t,j}}{\|\vect{R}_{obs} \vect{K}_{obs}^{-1} \vect{x}_{t,j}\|}
\end{equation}
Directly using rays in the observer camera frame would not work well in our scenario: the observer's fast and frequent head movements can cause severe high-frequency rotational variance, even when the wearer remains still. By leveraging $\vect{R}_{obs}$ to decouple ray and observer's ego-motion, $\hat{\vect{d}}_{t,j}$ provides a stable, rotation-invariant directional anchor.

\np{Depth scaling.} 
$\hat{\vect{d}}_{t,j}$ discards the spatial distance between the two subjects, which is critical for resolving scale ambiguities. We therefore scale $\hat{\vect{d}}_{t,j}$ by the Euclidean distance between observer and wearer head positions:
\begin{equation}
    \tilde{\vect{d}}_{t,j} = \hat{\vect{d}}_{t,j} \cdot \|\vect{p}_t^{w,\text{head}} - \vect{p}_t^{o,\text{head}}\|
\end{equation}
We then anchor this scaled vector to the observer's global head position to compute a proxy 3D endpoint position: 
\begin{equation}
    \vect{e}_{t,j} = \tilde{\vect{d}}_{t,j} + \vect{p}_t^{o,\text{head}}
\end{equation}
Geometrically, $\vect{e}_{t,j}$ lies precisely along the observer's line of sight, at a depth proportional to the inter-person distance.

\np{Ego-space canonicalization.}
Keeping these endpoints in the global frame makes the representation dependent on the wearer's global position and orientation: the same body pose performed at different locations would result in different representations, affecting generalization. We ensure a canonical representation by transforming each endpoint into the wearer's head-local coordinate frame:
\begin{equation}
    \vect{r}_{t,j} = \vect{R}_{w, head}^{-1}(\vect{e}_{t,j} - \vect{p}_t^{w,\text{head}})
\end{equation}
In this canonical frame, the relationship between proxy endpoints and ground-truth body pose is invariant to the wearer's global trajectory and camera rotation.
The concatenated representation $\vect{r}_t = [\vect{r}_{t,1}, \dots, \vect{r}_{t,K}] \in \mathbb{R}^{3K}$ serves as our robust geometric exocentric feature. In summary, this representation accounts for wearer--observer distance, factors out observer camera rotations, and removes dependence on the wearer's absolute
global position and orientation. We validate these design choices in \Cref{sec:ablation}.

\subsection{Learned Visibility Gating}
\label{sec:visibility_gating}

The reliability of the exocentric signal varies considerably over time: the wearer may be partially occluded, leave the observer's field of view, or be detected with poor accuracy at certain keypoints. Feeding noisy or absent rays into the fusion network without modulation would corrupt the prediction. 
Recent ray-based formulations (\eg, LAMP~\cite{yang2026lamp}) attach 2D detector
confidences to the lifted rays.
However, we find these confidences are not always reliable as visibility estimates under occlusion: erroneous keypoints may still receive high scores.
To address this, we learn a soft per-joint gating mechanism driven by the global semantic context of the observer's image. Specifically, from the same cropped region used for keypoint detection, we extract a DINOv3 CLS token $\vect{F}_{\text{CLS}} \in \mathbb{R}^{768}$ and map it to $K$ per-joint confidence scores through \textbf{ScoreNet}, which consists of a two-layer MLP ($768 \!\to\! 512 \!\to\! K$), followed by a sigmoid function $\sigma$:
\begin{equation}
    \vect{w} = \sigma(\text{MLP}(\vect{F}_{\text{CLS}})) \in [0,1]^{K},
\end{equation} Each canonicalized ray $\vect{r}_{t,j}$ from \Cref{sec:ray_canonicalization} is then element-wise scaled by its corresponding confidence:
\begin{equation}
    \hat{\vect{r}}_{t,j} = w_{t,j} \cdot \vect{r}_{t,j}, \quad j = 1, \dots, K.
\end{equation}
When the wearer is not well observable, the gate automatically suppresses the exocentric signal, encouraging the network to fall back on egocentric tracking.

We observed that a learned gating function can capture richer semantics from the holistic image context: for instance, it can recognize when the wearer is behind furniture even if the 2D detector still produces high-confidence but erroneous detections (\Cref{sec:ablation}).

\subsection{Ego-Exo View Aggregation}
\label{sec:fusion}

\np{Token construction.}
The egocentric signal and the coarse pose predicted by EgoNet (\Cref{sec:wearer_localization}) are concatenated to form the Ego Token, which is projected to a 512-dimensional embedding. The gated exocentric rays $\hat{\vect{r}}_{t,j}$ from \Cref{sec:visibility_gating} are likewise projected to the same dimension, yielding the Exo Token. Each frame thus produces a pair of tokens encoding complementary information: the Ego Token carries accurate but spatially sparse device tracking together with a full-body prior, while the Exo Token provides visually grounded full-body geometry.

\np{Spatial fusion.}
Per-frame Ego and Exo Tokens are arranged into a short sequence $[\vect{e}_t, \vect{o}_t]$ and processed by a Spatial Transformer Encoder. Self-attention allows the ego token to selectively attend to and absorb relevant information from the exocentric observation. We retain only the ego token's output $\vect{e}_t$ as the frame-level fused feature, enforcing an egocentric inductive bias: the egocentric signal serves as the primary representation, and the exocentric observation acts as an auxiliary enhancement. This ensures that even when the exocentric signal is entirely suppressed by the confidence gate (\Cref{sec:visibility_gating}), the network can still produce a reasonable prediction from the egocentric tracking alone. This design also scales to $N$ observers seamlessly by simply extending the token sequence to $[\vect{e}_t, \vect{o}_t^1, \dots, \vect{o}_t^N]$ without any architectural change.

\np{Temporal modeling and decoding.}
The fused features from all frames $\{\tilde{\vect{e}}_t\}_{t=1}^{T}$, augmented with learnable temporal positional embeddings, are passed through a Temporal Transformer Encoder that performs bidirectional self-attention over the full window of $T{=}96$ frames, capturing long-range motion dynamics. Each output frame is independently decoded by two lightweight MLP heads: one producing the root orientation $\hat{\boldsymbol{\theta}}_t^{w,\text{root}} \in \mathbb{R}^{6}$, and the other producing $J{=}21$ body joint rotations $\hat{\boldsymbol{\theta}}_t^{w,\text{body}} \in \mathbb{R}^{126}$, both in 6D rotation representation. The root translation is not predicted by the network; instead, we recover it analytically: given the known head world position $\vect{p}_t^{w,\text{head}}$ from the HMD and the predicted joint rotations $\hat{\boldsymbol{\theta}}_t$, we compute the head-to-root offset via forward kinematics and subtract it: $\vect{p}_t^{w,\text{root}} = \vect{p}_t^{w,\text{head}} - \text{FK}_{\text{head}}(\hat{\boldsymbol{\theta}}_t)$, where $\text{FK}_{\text{head}}$ returns the head joint position relative to the root in the body's local frame~\cite{jiang2022avatarposer,jiang2024egoposer, ma2024nymeria}.

\subsection{Training}
\label{sec:training}

\np{Two-stage training.}
We train the pipeline in two stages. In the first stage, EgoNet (\Cref{sec:wearer_localization}) is trained using only the egocentric signal, without any exocentric input. In the second stage, we freeze the EgoNet's coarse predictions and train the full ego-exo fusion network end-to-end, including the ray embedding, DINO confidence gate, Spatial Transformer, and Temporal Transformer. The DINOv3 backbone is kept frozen throughout; only the gating MLP is trained.

\np{Loss function.}
The training objective combines three complementary L1 losses:
\begin{equation}
    \mathcal{L} = \lambda_{\text{orient}}\,\mathcal{L}_{\text{orient}} + \lambda_{\text{rot}}\,\mathcal{L}_{\text{rot}} + \lambda_{\text{pos}}\,\mathcal{L}_{\text{pos}},
\end{equation}
where $\mathcal{L}_{\text{orient}}$ supervises the root orientation in 6D rotation space, $\mathcal{L}_{\text{rot}}$ supervises the $J{=}21$ body joint rotations in 6D space, and $\mathcal{L}_{\text{pos}}$ penalizes the L1 error on 3D joint positions obtained via SMPL forward kinematics from the predicted rotations. The rotation and position losses are complementary: rotation loss provides direct supervision in the rotation space, while position loss propagates gradients through the kinematic chain and penalizes error accumulation at distal joints. We set $\lambda_{\text{orient}} = 0.02$, $\lambda_{\text{rot}} = 1.0$, and $\lambda_{\text{pos}} = 1.0$.

        \begin{figure*}[t!]
    \centering
    \includegraphics[width=0.95\linewidth]{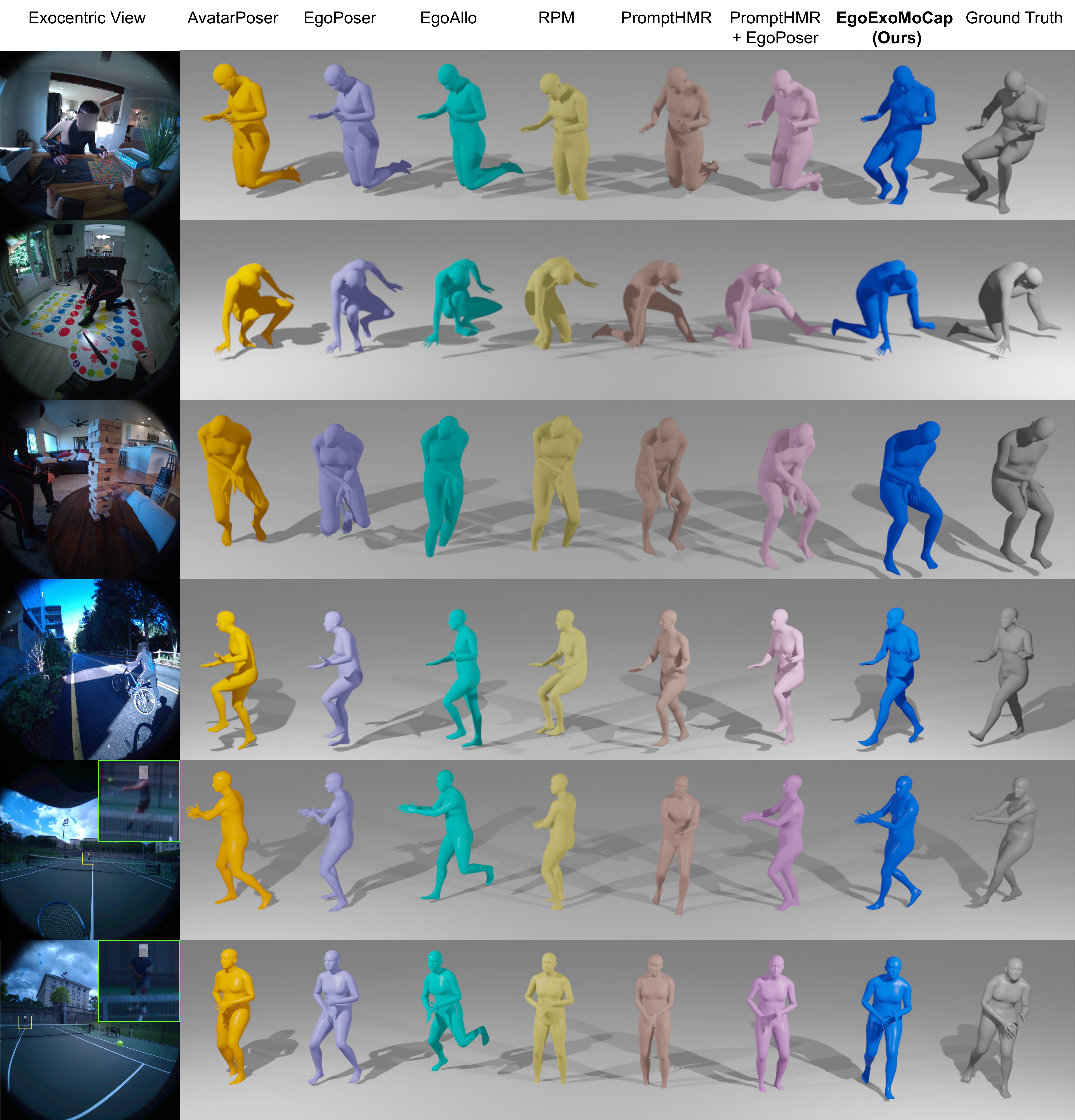}
    \caption{Qualitative comparison of \name{} versus baselines. The first four rows compare diverse activities from Nymeria, while the last two rows are drawn from the EgoHumans dataset and show two players playing tennis together.}
    \label{fig:qualitative-nymeria}
\end{figure*}
\section{Experiments}

\subsection{Experimental Protocol}

\np{Datasets.} We train and test on the large-scale \textbf{Nymeria}~\cite{ma2024nymeria} dataset, which pairs multi-modal egocentric streams from \emph{Aria} glasses~\cite{engel2023project} with ground-truth full-body motions obtained with the Xsens inertial system~\cite{xsens}. 
We use the SMPL representation provided by NymeriaPlus~\cite{detone2026nymeriaplus}, which was retargeted from the original ground-truth motion.
Nymeria features 300 hours of diverse indoor and outdoor activities. It also provides 6DoF wrist trajectories obtained by \emph{miniAria} wristbands.
We randomly split the dataset into 80\% training and 20\% test sets, ensuring no test-train overlap across subjects. 

To assess generalization, we additionally perform cross-dataset evaluation on \textbf{EgoHumans}~\cite{egohumans}, an outdoor multi-person dataset also captured with Aria glasses, featuring dynamic interactive activities such as fencing, basketball, and badminton. Ground-truth motions are obtained using a multi-view camera system, limiting the capture area but removing the requirement for subjects to wear an inertial suit as in Nymeria. Since EgoHumans sequences do not provide wrist tracking signals, we use the synthesized 6DoF tracking signals from ground-truth body parameters during evaluation.

\np{Sensor setup.} We evaluate our method under two tracking configurations based on the \textbf{wearer}'s device setup: (i)\textbf{ 3-point tracking}, where the wearer uses glasses plus two wrist-worn devices (or controllers) providing wrist trajectories; and (ii)\textbf{ 1-point tracking}, where the wearer uses only the glasses, providing head trajectory alone. In both setups, the \textbf{observer} wears only the glasses.

\np{Metrics.} We consider both positional and physical-plausibility evaluation metrics~\cite{jiang2022avatarposer,jiang2024egoposer,yi2025estimating,barquero2025sparse}:
    Mean Per-Joint Position Error (\textbf{MPJPE}, cm), alongside Upper-body (\textbf{U-PE}) and Lower-body (\textbf{L-PE}) position errors; 
    Mean Per-Joint Velocity Error (\textbf{MPJVE}, cm/s), measuring the difference between predicted and ground-truth joint velocities; 
    Motion \textbf{Jitter} ($10^2$ m/s$^3$), calculated as the mean magnitude of the third derivative of position (jerk).
    
\np{Baselines.}
We benchmark our method against several state-of-the-art pose estimation approaches, considering both egocentric and exocentric ones, using their open-sourced code.
As egocentric baselines, we include regression models, such as AvatarPoser~\cite{jiang2022avatarposer} and EgoPoser~\cite{jiang2024egoposer}, alongside Diffusion-based frameworks like EgoAllo~\cite{yi2025estimating} and RPM~\cite{barquero2025sparse}. We retrain and evaluate them under 1-point and 3-point settings on the same train-test split as ours. We partition the full sequence into non-overlapping $T$-frame segments during testing. 
Note that, for the experiments, we improved baselines ~\cite{jiang2022avatarposer} and ~\cite{jiang2024egoposer} by predicting a sequence instead of just the last frame, which provides better results than their online tracking designs.

We consider PromptHMR~\cite{wang2025prompthmr} as an exocentric baseline.
As its training code is not available, for a fairer comparison, we adapt the original model by freezing it and adding to it a Transformer-based output layer, trained on Nymeria.
We feed PromptHMR with ground-truth observer camera parameters and ground-truth wearer global position and orientation. 
We report both the original PromptHMR output and a finetuned variant, PromptHMR-Finetuned, where the original model is frozen and followed by a Transformer-based output layer trained on Nymeria.
Furthermore, we implement a custom baseline that integrates the outputs of PromptHMR and EgoPoser through an additional Transformer network.
Following~\cite{barquero2025sparse}, our evaluation setup accounts for individual user dimensions, rather than assuming a universal average body shape~\cite{jiang2022avatarposer, du2023avatars}, using SMPL identity ground-truth parameters. 

\subsection{Results}

\Cref{tab:comparison} summarizes the quantitative results on Nymeria. Visual comparisons are provided in~\Cref{fig:qualitative-nymeria}. Across all metrics and both tracking setups, our method outperforms the baselines. The only exception is Jitter, where RPM achieves the best performance; we hypothesize this is due to its PCAF module~\cite{barquero2025sparse}, which balances smoothness with a potentially minor adherence to input signals. 

In general, egocentric methods can return plausible motions (often exhibiting low jitter) but cannot faithfully reconstruct invisible parts (\eg, kneeling or sitting poses). PromptHMR, even if fed with ground-truth root position and rotation, struggles with extreme occlusions (\eg, intervals in which the wearer is not visible at all) and observer's HMD camera motion.
The combination of PromptHMR and EgoPoser in our ego-exo baseline shows the benefits of combining both sources of information, reporting low MPJPE for both upper and lower body.
However, its ``naive'' fusion of ego- and exocentric features results in decreased accuracy and reduced smoothness (higher MPJVE and Jitter). Early fusion of 2D and 3D features, as performed in our method, helps increase robustness: the estimate does not rely excessively on the exocentric input when this is noisy and unreliable (see \eg first row in~\Cref{fig:qualitative-nymeria}, where naive fusion does not recover from the inaccurate PromptHMR estimate).

\begin{figure*}[t!]
    \centering
    \includegraphics[width=\linewidth]{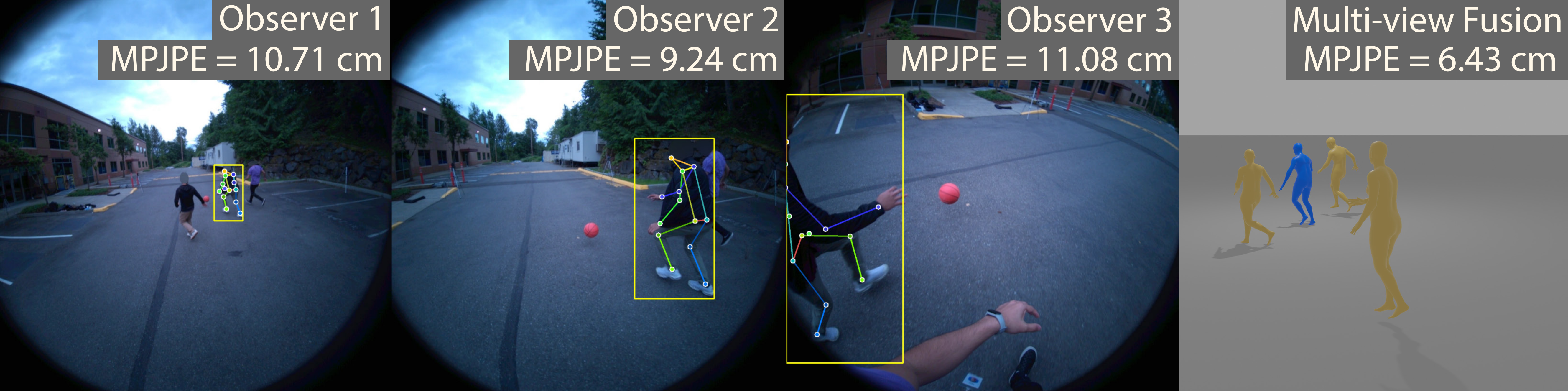}
    \caption{\textbf{Multi-observer fusion.} Egocentric tracking with a single exocentric view (Observers 1-3) may struggle with partial observations and severe body truncation, yielding MPJPEs around 9 to 11 cm. Our approach effectively aggregates these arbitrary views, dropping the final 3D pose error to 6.43 cm (right).} 
    \label{fig:multi-view}
\end{figure*}

\cref{tab:egohumans-comparison} reports quantitative results on EgoHumans, confirming the trend observed in Nymeria.
We observe how scaling our approach to multiple observers can bring additional benefit. \cref{fig:multi-view} shows a typical scenario: single observers may only have partial or far-away views of the wearer, leading to suboptimal pose estimates; combining their views, accuracy significantly improves.
We also compare against a baseline using multi-view triangulated keypoints as exo tokens instead of fused 2D and DINO features: the baseline does not adequately account for confidence scores associated with different views and exhibits worse results. 
These results suggest the potential of distributed HMD-based setups for in-the-wild motion capture. 

\begin{table*}[t!]
\centering
\setlength{\tabcolsep}{2pt}
\caption{\textbf{Quantitative evaluation on Nymeria.}
We compare ego- and exocentric methods under the 1-point and 3-point setups.
Best results highlighted in \textbf{boldface}.}
\begin{adjustbox}{width=\textwidth}
\begin{tabular}{lcrrrrrrrrrr}
\toprule
\multirow{2}{*}{Methods}  &\multirow{2}{*}{~~~Input Modality~~~}&\multicolumn{5}{c}{Three-Point Tracking} &  \multicolumn{5}{c}{One-Point Tracking} \\
        \cmidrule(lr){3-7} \cmidrule(lr){8-12} 
  && MPJPE~ & Upper~ &  Lower~& MPJVE~&Jitter~ & MPJPE~ & Upper~ &  Lower~&MPJVE~&Jitter~ \\ \midrule
AvatarPoser~\cite{jiang2022avatarposer} &Ego& 8.16&3.82&14.43&13.62&2.20& 12.38& 9.22& 16.94& 18.90& 4.07\\
 EgoPoser~\cite{jiang2024egoposer}
 &Ego& 7.74& 3.44& 13.96& 13.94& 2.43& 11.86& 9.08& 15.87& 20.26& 3.43\\
 EgoAllo~\cite{yi2025estimating} &Ego& 8.77& 3.18& 16.84& 12.81& 1.72& 12.53& 8.27& 18.68& 20.17& 1.96\\
 RPM~\cite{barquero2025sparse} &Ego& 12.19& 3.50& 24.74& 17.62& \textbf{1.29}& 16.83& 9.39& 27.58& 18.93& \textbf{1.15}\\
 PromptHMR~\cite{wang2025prompthmr} &Exo& 13.48&8.88&20.13&26.78&5.17&13.48&8.88&20.13&26.78&5.17\\  
 PromptHMR-Finetuned &Exo& 10.65& 7.63& 15.01& 23.55& 4.98&10.65& 7.63& 15.01& 23.55& 4.98\\  
 PromptHMR+EgoPoser  &EgoExo& 6.47& 3.41& 10.90& 17.49& 3.08& 9.03& 6.74& 12.34& 17.52& 3.17\\
 \textbf{\name{} (Ours)}
  &EgoExo& \textbf{5.72}&\textbf{2.72}&\textbf{10.05}&\textbf{12.77}&2.16&\textbf{ 8.28}& \textbf{6.10}&\textbf{ 11.44}& \textbf{17.23}& 2.47\\ 
 \bottomrule
 \end{tabular}
 \end{adjustbox}
\label{tab:comparison}
\end{table*}

\begin{table*}[t!]
\centering
\setlength{\tabcolsep}{2pt}
\caption{\textbf{Quantitative evaluation on EgoHumans.}
We compare ego- and exocentric methods under the 1-point and 3-point setups. We also evaluate our method in the multi-observer setup, on the EgoHumans subset(*) providing multi-observer streams.
Best results highlighted in \textbf{boldface}.}
\begin{adjustbox}{width=\textwidth}
\begin{tabular}{lcrrrrrrrrrr}
\toprule
\multirow{2}{*}{Methods}  &\multirow{2}{*}{~~~Input Modality~~~}&\multicolumn{5}{c}{Three-Point Tracking} &  \multicolumn{5}{c}{One-Point Tracking} \\
        \cmidrule(lr){3-7} \cmidrule(lr){8-12} 
  && MPJPE~ & Upper~ &  Lower~& MPJVE~&Jitter~ & MPJPE~ & Upper~ &  Lower~&MPJVE~&Jitter~ \\ \midrule
AvatarPoser~\cite{jiang2022avatarposer} &Ego& 9.60&4.34&17.18&32.94&3.08& 14.17& 10.38& 19.64& 54.60& 3.84\\
 EgoPoser~\cite{jiang2024egoposer}
 &Ego& 9.03& 3.87& 16.48& 33.12& 3.55& 13.96& 10.49& 18.96& 57.04& 5.05\\
  EgoAllo~\cite{yi2025estimating} &Ego& 10.19& 3.82& 19.39& 36.82& 1.94& 14.38& 11.27& 18.87& 51.77& 1.85\\
 RPM~\cite{barquero2025sparse} &Ego& 10.69& 4.79& 19.21& 31.87& \textbf{0.85}&13.70& 11.03& 17.56& 43.52& \textbf{1.14}\\
 PromptHMR~\cite{wang2025prompthmr} &Exo& 18.40&12.91&26.34&58.63&10.50&18.40&12.91&26.34&58.63&10.50\\  
 PromptHMR-Finetuned &Exo& 18.15& 11.79& 27.34& 46.84& 5.17&18.15& 11.79& 27.34& 46.84& 5.17\\  
 PromptHMR+EgoPoser  &EgoExo& 8.62& 4.24& 14.94& 34.01& 2.79& 13.27& 8.96& 19.50& 41.88& 2.71\\
 \textbf{\name{} (Ours)} &EgoExo&\textbf{7.62}& \textbf{3.45}&\textbf{13.64}&\textbf{30.94}&1.83&\textbf{11.54}&\textbf{8.18}& \textbf{16.39}& \textbf{41.04}& 2.07\\ 
  \midrule
  EgoExo-single-observer$^*$ & EgoExo& 8.80& 7.00&11.40&47.23&3.59& 13.70& 12.35& 15.64& 50.32& 3.01\\
 EgoExo-triangulation$^*$& Ego-Multi-Exo& 8.49& 6.58& 11.25& 44.38& 3.23& 13.34& 11.83& 15.52& 49.04&2.90\\ 
  EgoExo-multi-observer$^*$ (Ours)& Ego-Multi-Exo& \textbf{7.11}& \textbf{5.89}&\textbf{8.87}&\textbf{37.31}&\textbf{2.58}&\textbf{ 10.48}& \textbf{9.59}& \textbf{11.77}& \textbf{45.50}& \textbf{2.83}\\ 
 \bottomrule
 \end{tabular}
 \end{adjustbox}
\label{tab:egohumans-comparison}
\end{table*}

\begin{figure*}[t!]
    \centering
    \includegraphics[width=\linewidth]{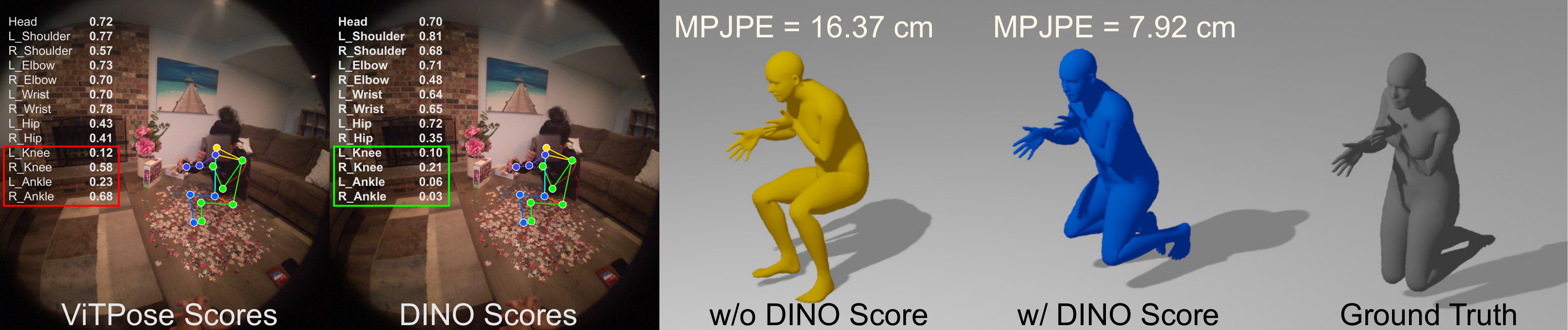}
    \caption{\textbf{ViTPose vs DINOv3 scores.} Learning DINO-based scores significantly improves robustness against 2D detections, typically observed under occlusion.}
    \label{fig:dino-ablation}
\end{figure*}

\subsection{Ablation Studies}
\label{sec:ablation}
 We ablate the components of our approach on the Nymeria test set in~\Cref{tab:ablation}.

\np{Ego+exo, wearer localization.} Removing either the egocentric input signals (w/o Ego) or the exocentric images (w/o Exo) leads, as expected, to a significant metrics drop.
Using a bounding box detector (YOLO~\cite{yolo}) instead of EgoNet predictions (w/o Ego BBX) worsens metrics. Detectors like YOLO typically struggle in the presence of strong body occlusions.

\np{Ray-based pose representation.} Omitting depth scaling (w/o Depth Scaling) degrades spatial reasoning and therefore accuracy.
Not converting rays from the observer’s frame to the wearer’s frame (Ray in World-Space, Ray in Exo-Space) also negatively impacts performance. This transformation helps factor out the observer's head motion, ensuring better generalization.

\np{Learned gating.} To evaluate the effectiveness of learned DINO scores, we remove them, keeping the original rays without gating (w/o DINO Score), replace them with ViT-provided confidences (w/ ViT score), and mask them out when ViT-provided confidence scores are smaller than 0.2 (w/ Masking). As \cref{fig:dino-ablation} exemplifies, DINO scores help in the presence of noisy 2D keypoints, \eg when the subject is occluded by furniture (recognized as not belonging to the body by DINO). Using ViT Score slightly improves some metrics over no gating, but remains worse than our learned DINO-based gating, especially for lower-body accuracy, given its unreliability in occluded scenarios.

\np{Robustness to EgoNet.} To further assess the dependence on EgoNet, we perturb its estimated joint positions with Gaussian noise of $\sigma=1/2/5/10$ cm: the final full-body MPJPE increases by only
0.005/0.02/0.14/0.54~cm, respectively, indicating graceful degradation and limited sensitivity to EgoNet's coarse pose estimates.

\begin{table}[t!]
    \centering
    \caption{Ablation study on Nymeria. Best results highlighted in \textbf{boldface}.}
\setlength{\tabcolsep}{2pt}
\begin{adjustbox}{width=0.48\textwidth}
    \begin{tabular}{llrrrrr}
\toprule
          &Methods& MPJPE & Upper & Lower & MPJVE & Jitter \\
         \midrule
 \emph{Default} &EgoExoMoCap&\textbf{5.72}& \textbf{2.72}&\textbf{10.05}&\textbf{12.77}&\textbf{2.16}\\
         \midrule
\multirow{3}{*}{\emph{Ego-Exo}}          &w/o Ego& 11.45&8.00&16.43& 34.40& 6.55\\
          &w/o Exo& 7.53& 3.38& 13.53& 13.78& 2.19\\
          &w/o Ego BBX& 6.43&3.05&11.31&15.54&2.80\\
          \midrule
\multirow{3}{*}{\emph{Learned Gating}}          &w/o DINO Score & 6.30& 3.00& 11.08& 14.42& 2.38\\
          &w/ ViT Score & 6.29& 2.92& 11.16& 13.91& 2.23\\
 & w/ Masking& 6.32& 2.99& 11.14& 15.31&2.68\\
          \midrule
\multirow{4}{*}{\emph{Ray Representation}}  
          &w/ Ray in World-Space & 6.99& 3.15& 12.54& 13.87& 2.44\\
          &w/ Ray in Exo-Space  & 6.14& 2.92& 10.78& 13.22& 2.21\\
          &w/o Depth Scaling& 6.26& 2.99& 10.97& 13.26& 2.19\\
          &w/o Lifting &6.26& 2.97& 11.00& 13.81& 2.33\\

  \bottomrule
    \end{tabular}
    \end{adjustbox}
    \label{tab:ablation}
\end{table}

\subsection{Discussion}
Our approach focuses on portable, lightweight human motion capture using a distributed HMD setup. We assume cameras are calibrated and synchronized by recent Aria tooling~\cite{aria2}. While we focus on in-the-wild ground-truth motion acquisition rather than online and real-time tracking, exploring real-time applications is an exciting direction for future work when hardware and tooling can support it.

Currently, we mainly focus on motion reconstruction and assume subject shape is provided when calculating the joint positions; it could potentially be estimated leveraging images or sensor-based calibrations~\cite{pujades2019virtual}.
We observed failures when the wearer is largely occluded by another subject -- since the scenario may confuse DINO-based visibility scores. Long out-of-view intervals or persistently unreliable exo signals (\eg, under heavy occlusion) might lower accuracy, especially for the lower body. We found learned gating helpful in assessing signal reliability in these scenarios: performance falls towards the ego-only method, leading to still plausible (but less faithful) motions. Better modeling of physical plausibility (foot-ground contact, body-self penetrations), together with better incorporation of scene context, could further enhance motion realism. 
        \section{Conclusion}
\label{sec:conclusion}
We presented \name{}, a novel scalable approach effectively combining egocentric and exocentric tracking for in-the-wild human motion capture.
Unlike traditional motion capture systems that rely on extensive hardware and complex setups, our method only leverages a set of HMDs, proposing a lightweight distributed solution for flexible and unobtrusive tracking in real-world environments. \name{} requires just two subjects, each wearing a pair of glasses, and can naturally scale to multi-subject setups. Extensive experiments on two in-the-wild datasets show that the approach performs favorably with respect to egocentric and exocentric baselines, handling challenging scenarios like occlusions and out-of-view motions.

Future work should explore the combination of further multi-modal streams (\eg, stereo cameras mounted on HMDs~\cite{engel2023project} or multiple body-worn sensors~\cite{armani2024accurate}) to provide richer motion cues. 
Tracklet association mechanisms from exocentric people tracking systems~\cite{yang2026lamp} could further extend EgoExoMoCap to crowded scenarios with non-HMD participants.
Furthermore, leveraging the 3D scene reconstruction capabilities of HMDs~\cite{ma2024nymeria} could lead to more accurate reconstruction of human-scene interactions.
\clearpage

        {
          \small
            \bibliographystyle{ieeenat_fullname}
              \bibliography{main}

@String(CVPR= {IEEE Conf. Comput. Vis. Pattern Recog.})

@String(ICCV= {Int. Conf. Comput. Vis.})

@String(ECCV= {Eur. Conf. Comput. Vis.})

@String(TOG= {ACM Trans. Graph.})

@String(CVPR  = {CVPR})

@String(ICCV  = {ICCV})

@String(ECCV  = {ECCV})

@String(TOG   = {ACM TOG})

@misc{aria2,
  title = {Aria Gen2 glasses},
  howpublished = {\url{https://ai.meta.com/blog/aria-gen-2-research-glasses-under-the-hood-reality-labs/}},
  note = {Accessed: 2026-02-28}
}

@inproceedings{dittadi2021full,
  title={Full-Body Motion From a Single Head-Mounted Device: Generating SMPL Poses From Partial Observations},
  author={Dittadi, Andrea and Dziadzio, Sebastian and Cosker, Darren and Lundell, Ben and Cashman, Thomas J and Shotton, Jamie},
  booktitle={Proceedings of the IEEE/CVF International Conference on Computer Vision},
  pages={11687--11697},
  year={2021}
}

@inproceedings{chi2024m2d2m,
  title={M2d2m: Multi-motion generation from text with discrete diffusion models},
  author={Chi, Seunggeun and Chi, Hyung-gun and Ma, Hengbo and Agarwal, Nakul and Siddiqui, Faizan and Ramani, Karthik and Lee, Kwonjoon},
  booktitle={European conference on computer vision},
  pages={18--36},
  year={2024},
  organization={Springer}
}

@inproceedings{von2017sparse,
  title={Sparse inertial poser: Automatic 3d human pose estimation from sparse imus},
  author={Von Marcard, Timo and Rosenhahn, Bodo and Black, Michael J and Pons-Moll, Gerard},
  booktitle={Computer graphics forum},
  volume={36},
  number={2},
  pages={349--360},
  year={2017},
  organization={Wiley Online Library}
}

@inproceedings{hollidt2024egosim,
author = {Hollidt, Dominik and Streli, Paul and Jiang, Jiaxi and Haghighi, Yasaman and Qian, Changlin and Liu, Xintong and Holz, Christian},
title = {{EgoSim: an egocentric multi-view simulator and real dataset for body-worn cameras during motion and activity}},
year = {2024},
booktitle = {Advances in Neural Information Processing Systems}
}

@article{loper2015smpl,
  title={SMPL: A skinned multi-person linear model},
  author={Loper, Matthew and Mahmood, Naureen and Romero, Javier and Pons-Moll, Gerard and Black, Michael J},
  journal={ACM transactions on graphics (TOG)},
  volume={34},
  number={6},
  pages={1--16},
  year={2015},
  publisher={ACM New York, NY, USA}
}

@article{tolstikhin2021mlp,
  title={Mlp-mixer: An all-mlp architecture for vision},
  author={Tolstikhin, Ilya O and Houlsby, Neil and Kolesnikov, Alexander and Beyer, Lucas and Zhai, Xiaohua and Unterthiner, Thomas and Yung, Jessica and Steiner, Andreas and Keysers, Daniel and Uszkoreit, Jakob and others},
  journal={Advances in neural information processing systems},
  volume={34},
  pages={24261--24272},
  year={2021}
}

@article{tolani2000real,
  title={Real-time inverse kinematics techniques for anthropomorphic limbs},
  author={Tolani, Deepak and Goswami, Ambarish and Badler, Norman I},
  journal={Graphical models},
  volume={62},
  number={5},
  pages={353--388},
  year={2000},
  publisher={Elsevier}
}

@inproceedings{cho2021camera,
  title={Camera distortion-aware 3d human pose estimation in video with optimization-based meta-learning},
  author={Cho, Hanbyel and Cho, Yooshin and Yu, Jaemyung and Kim, Junmo},
  booktitle={Proceedings of the IEEE/CVF international conference on computer vision},
  pages={11169--11178},
  year={2021}
}

@article{vaswani2017attention,
  title={Attention is all you need},
  author={Vaswani, Ashish and Shazeer, Noam and Parmar, Niki and Uszkoreit, Jakob and Jones, Llion and Gomez, Aidan N and Kaiser, {\L}ukasz and Polosukhin, Illia},
  journal={Advances in Neural Information Processing Systems},
  volume={30},
  year={2017}
}

@InProceedings{Luo_2024_CVPR,
    author    = {Luo, Zhengyi and Cao, Jinkun and Khirodkar, Rawal and Winkler, Alexander and Kitani, Kris and Xu, Weipeng},
    title     = {Real-Time Simulated Avatar from Head-Mounted Sensors},
    booktitle = {Proceedings of the IEEE/CVF Conference on Computer Vision and Pattern Recognition (CVPR)},
    year      = {2024},
    pages     = {571-581}
}

@inproceedings{zhou2019continuity,
  title={On the continuity of rotation representations in neural networks},
  author={Zhou, Yi and Barnes, Connelly and Lu, Jingwan and Yang, Jimei and Li, Hao},
  booktitle={Proceedings of the IEEE/CVF Conference on Computer Vision and Pattern Recognition},
  pages={5745--5753},
  year={2019}
}

@InProceedings{dynaip2024,
    author    = {Zhang, Yu and Xia, Songpengcheng and Chu, Lei and Yang, Jiarui and Wu, Qi and Pei, Ling},
    title     = {Dynamic Inertial Poser (DynaIP): Part-Based Motion Dynamics Learning for Enhanced Human Pose Estimation with Sparse Inertial Sensors},
    booktitle = {Proceedings of the IEEE/CVF Conference on Computer Vision and Pattern Recognition (CVPR)},
    year      = {2024}
}

@article{dhamanaskar2023enhancing,
  title={Enhancing egocentric 3d pose estimation with third person views},
  author={Dhamanaskar, Ameya and Dimiccoli, Mariella and Corona, Enric and Pumarola, Albert and Moreno-Noguer, Francesc},
  journal={Pattern Recognition},
  volume={138},
  pages={109358},
  year={2023},
  publisher={Elsevier}
}

@inproceedings{zhang2023probabilistic,
  title={Probabilistic human mesh recovery in 3d scenes from egocentric views},
  author={Zhang, Siwei and Ma, Qianli and Zhang, Yan and Aliakbarian, Sadegh and Cosker, Darren and Tang, Siyu},
  booktitle={Proceedings of the IEEE/CVF International Conference on Computer Vision},
  pages={7989--8000},
  year={2023}
}

@inproceedings{kareer2025egomimic,
  title={Egomimic: Scaling imitation learning via egocentric video},
  author={Kareer, Simar and Patel, Dhruv and Punamiya, Ryan and Mathur, Pranay and Cheng, Shuo and Wang, Chen and Hoffman, Judy and Xu, Danfei},
  booktitle={2025 IEEE International Conference on Robotics and Automation (ICRA)},
  pages={13226--13233},
  year={2025},
  organization={IEEE}
}

@article{hoque2025egodex,
  title={Egodex: Learning dexterous manipulation from large-scale egocentric video},
  author={Hoque, Ryan and Huang, Peide and Yoon, David J and Sivapurapu, Mouli and Zhang, Jian},
  journal={arXiv preprint arXiv:2505.11709},
  year={2025}
}

@misc{yang2025egovlalearningvisionlanguageactionmodels,
title={EgoVLA: Learning Vision-Language-Action Models from Egocentric Human Videos}, 
author={Ruihan Yang and Qinxi Yu and Yecheng Wu and Rui Yan and Borui Li and An-Chieh Cheng and Xueyan Zou and Yunhao Fang and Hongxu Yin and Sifei Liu and Song Han and Yao Lu and Xiaolong Wang},
year={2025},
eprint={2507.12440},
archivePrefix={arXiv},
primaryClass={cs.RO}
}

@article{punamiya2026egoverse,
  title={Egoverse: An egocentric human dataset for robot learning from around the world},
  author={Punamiya, Ryan and Kareer, Simar and Liu, Zeyi and Citron, Josh and Qiu, Ri-Zhao and Cai, Xiongyi and Gavryushin, Alexey and Chen, Jiaqi and Liconti, Davide and Zhu, Lawrence Y and others},
  journal={arXiv preprint arXiv:2604.07607},
  year={2026}
}

@inproceedings{shi2025caring,
  title={Caring-ai: Towards authoring context-aware augmented reality instruction through generative artificial intelligence},
  author={Shi, Jingyu and Jain, Rahul and Chi, Seunggeun and Doh, Hyungjun and Chi, Hyung-gun and Quinn, Alexander J and Ramani, Karthik},
  booktitle={Proceedings of the 2025 CHI conference on human factors in computing systems},
  pages={1--23},
  year={2025}
}

@article{li2023object,
  title={Object motion guided human motion synthesis},
  author={Li, Jiaman and Wu, Jiajun and Liu, C Karen},
  journal={ACM Transactions on Graphics (TOG)},
  volume={42},
  number={6},
  pages={1--11},
  year={2023},
  publisher={ACM New York, NY, USA}
}

@inproceedings{araujo2023circle,
  title={Circle: Capture in rich contextual environments},
  author={Ara{\'u}jo, Joao Pedro and Li, Jiaman and Vetrivel, Karthik and Agarwal, Rishi and Wu, Jiajun and Gopinath, Deepak and Clegg, Alexander William and Liu, Karen},
  booktitle={Proceedings of the IEEE/CVF Conference on Computer Vision and Pattern Recognition},
  pages={21211--21221},
  year={2023}
}

@article{shi2026egohumanoid,
  title={Egohumanoid: Unlocking in-the-wild loco-manipulation with robot-free egocentric demonstration},
  author={Shi, Modi and Peng, Shijia and Chen, Jin and Jiang, Haoran and Li, Yinghui and Huang, Di and Luo, Ping and Li, Hongyang and Chen, Li},
  journal={arXiv preprint arXiv:2602.10106},
  year={2026}
}

@inproceedings{wang2022estimating,
  title={Estimating egocentric 3d human pose in the wild with external weak supervision},
  author={Wang, Jian and Liu, Lingjie and Xu, Weipeng and Sarkar, Kripasindhu and Luvizon, Diogo and Theobalt, Christian},
  booktitle={Proceedings of the IEEE/CVF Conference on Computer Vision and Pattern Recognition},
  pages={13157--13166},
  year={2022}
}

@article{detone2026nymeriaplus,
  title={NymeriaPlus: Enriching Nymeria Dataset with Additional Annotations and Data},
  author={DeTone, Daniel and Bogo, Federica and Le, Eric-Tuan and Frost, Duncan and Straub, Julian and Siddiqui, Yawar and Ye, Yuting and Engel, Jakob and Newcombe, Richard and Ma, Lingni},
  journal={arXiv preprint arXiv:2603.18496},
  year={2026}
}

@inproceedings{yang2026lamp,
  title     = {{LAMP}: Localization Aware Multi-camera People Tracking in Metric {3D} World},
  author    = {Yang, Nan and Straub, Julian and Zhang, Fan and Newcombe, Richard and Engel, Jakob and Ma, Lingni},
  booktitle = {Proceedings of the IEEE/CVF Conference on Computer Vision and Pattern Recognition (CVPR)},
  year      = {2026}
}

@inproceedings{aliakbarian2022flag,
  title={Flag: Flow-based 3d avatar generation from sparse observations},
  author={Aliakbarian, Sadegh and Cameron, Pashmina and Bogo, Federica and Fitzgibbon, Andrew and Cashman, Thomas J},
  booktitle={Proceedings of the IEEE/CVF Conference on Computer Vision and Pattern Recognition},
  pages={13253--13262},
  year={2022}
}

@inproceedings{aliakbarian2023nemo,
  title={{HMD-Nemo}: Online 3D avatar motion generation from sparse observation},
  author={Aliakbarian, Sadegh and Saleh, Fatemeh and Collier, David and Cameron, Pashmina and Cosker, Darren},
  booktitle={Proceedings of the IEEE/CVF Conference on Computer Vision and Pattern Recognition},
  pages={9622--9631},
  year={2023}
}

@inproceedings{jiang2022avatarposer,
  title={{Avatarposer: Articulated full-body pose tracking from sparse motion sensing}},
  author={Jiang, Jiaxi and Streli, Paul and Qiu, Huajian and Fender, Andreas and Laich, Larissa and Snape, Patrick and Holz, Christian},
  booktitle={Proceedings of the European Conference on Computer Vision (ECCV)},
  year={2022}
}

@inproceedings{winkler2022questsim,
  title={QuestSim: Human Motion Tracking from Sparse Sensors with Simulated Avatars},
  author={Winkler, Alexander and Won, Jungdam and Ye, Yuting},
  booktitle={SIGGRAPH Asia 2022 Conference Papers},
  year={2022}
}

@inproceedings{grauman2024ego,
  title={Ego-exo4d: Understanding skilled human activity from first-and third-person perspectives},
  author={Grauman, Kristen and Westbury, Andrew and Torresani, Lorenzo and Kitani, Kris and Malik, Jitendra and Afouras, Triantafyllos and Ashutosh, Kumar and Baiyya, Vijay and Bansal, Siddhant and Boote, Bikram and others},
  booktitle={Proceedings of the IEEE/CVF Conference on Computer Vision and Pattern Recognition},
  pages={19383--19400},
  year={2024}
}

@inproceedings{lee2023questenvsim,
  title={{QuestEnvSim: Environment-Aware Simulated Motion Tracking from Sparse Sensors}},
  author={Lee, Sunmin and Starke, Sebastian and Ye, Yuting and Won, Jungdam and Winkler, Alexander},
  booktitle = {ACM SIGGRAPH 2023 Conference Proceedings},
  year={2023}
}

@inproceedings{du2023avatars,
  title={Avatars Grow Legs: Generating Smooth Human Motion
from Sparse Tracking Inputs with Diffusion Model},
  author={Du, Yuming and Kips, Robin and Pumarola, Albert and Starke, Sebastian and Thabet, Ali and Sanakoyeu, Artsiom},
  booktitle={Proceedings of the IEEE/CVF Conference on Computer Vision and Pattern Recognition},
  year={2023}
}

@article{huang2018deep,
  title={Deep inertial poser: Learning to reconstruct human pose from sparse inertial measurements in real time},
  author={Huang, Yinghao and Kaufmann, Manuel and Aksan, Emre and Black, Michael J and Hilliges, Otmar and Pons-Moll, Gerard},
  journal={ACM Transactions on Graphics (TOG)},
  volume={37},
  number={6},
  pages={1--15},
  year={2018},
  publisher={ACM New York, NY, USA}
}

@inproceedings{feng2023sage,
  author    = {Feng, Han and Ma, Wenchao and Gao, Quankai and Zheng, Xianwei and Xue, Nan and Xu, Huijuan},
  title     = {Stratified Avatar Generation from Sparse Observations},
  booktitle   = {Proceedings of the IEEE conference on computer vision and pattern recognition},
pages={153--163},
  year      = {2024},
}

@article{chi2024estimating,
  title={Estimating Ego-Body Pose from Doubly Sparse Egocentric Video Data},
  author={Chi, Seunggeun and Huang, Pin-Hao and Sachdeva, Enna and Ma, Hengbo and Ramani, Karthik and Lee, Kwonjoon},
  journal={Advances in neural information processing systems},
  year={2024}
}

@inproceedings{zhang2022egobody,
  title={Egobody: Human body shape and motion of interacting people from head-mounted devices},
  author={Zhang, Siwei and Ma, Qianli and Zhang, Yan and Qian, Zhiyin and Kwon, Taein and Pollefeys, Marc and Bogo, Federica and Tang, Siyu},
  booktitle={Proceedings of the European Conference on Computer Vision (ECCV)},
  pages={180--200},
  year={2022}
}

@inproceedings{li2023ego,
  title={Ego-Body Pose Estimation via Ego-Head Pose Estimation},
  author={Li, Jiaman and Liu, Karen and Wu, Jiajun},
  booktitle={Proceedings of the IEEE/CVF Conference on Computer Vision and Pattern Recognition},
  pages={17142--17151},
  year={2023}
}

@inproceedings{
  zheng2023realistic,
  title={Realistic Full-Body Tracking from Sparse Observations via Joint-Level Modeling},
  author={Zheng, Xiaozheng and Zhuo Su and Wen, Chao and Xue, Zhou and Xiaojie Jin},
  booktitle={Proceedings of the IEEE/CVF international conference on computer vision},
  year={2023}
}

@misc{xsens,
  title = {Movella XSens},
  howpublished = {\url{https://www.movella.com/motion-capture/xsens-link-specifications}},
  note = {Accessed: 2026-02-28}
}

@inproceedings{von2018recovering,
  title={Recovering accurate 3d human pose in the wild using imus and a moving camera},
  author={Von Marcard, Timo and Henschel, Roberto and Black, Michael J and Rosenhahn, Bodo and Pons-Moll, Gerard},
  booktitle={Proceedings of the European conference on computer vision (ECCV)},
  pages={601--617},
  year={2018}
}

@inproceedings{dong2024,
    author = {Dong, Kun and Xue, Jian and Niu, Zehai and Lan, Xing and Lu, Ke and Liu, Qingyuan and Qin, Xiaoyu},
    title = {Realistic Full-Body Motion Generation from Sparse Tracking with State Space Model},
    year = {2024},
    publisher = {Association for Computing Machinery},
    booktitle = {Proceedings of the 32nd ACM International Conference on Multimedia},
    pages = {4024–4033},
}

@inproceedings{pan2023fusingmono,
author = {Pan, Shaohua and Ma, Qi and Yi, Xinyu and Hu, Weifeng and Wang, Xiong and Zhou, Xingkang and Li, Jijunnan and Xu, Feng},
title = {Fusing Monocular Images and Sparse IMU Signals for Real-Time Human Motion Capture},
year = {2023},
booktitle = {SIGGRAPH Asia 2023 Conference Papers},
}

@inproceedings{mollyn2023imuposer,
author = {Mollyn, Vimal and Arakawa, Riku and Goel, Mayank and Harrison, Chris and Ahuja, Karan},
title = {{IMUPoser: Full-Body Pose Estimation Using IMUs in Phones, Watches, and Earbuds}},
year = {2023},
booktitle = {Proceedings of the 2023 CHI Conference on Human Factors in Computing Systems}
}

@inproceedings{UIP,
  author = {Armani, Rayan and Qian, Changlin and Jiang, Jiaxi and Holz, Christian},
  title = {{Ultra Inertial Poser}: Scalable Motion Capture and Tracking from Sparse Inertial Sensors and Ultra-Wideband Ranging},
  year = {2024},
  booktitle = {ACM SIGGRAPH 2024 Conference Papers}
}

@INPROCEEDINGS{armani2024accurate,
  author={Armani, Rayan and Holz, Christian},
  booktitle={2024 IEEE/RSJ International Conference on Intelligent Robots and Systems (IROS)}, 
  title={Accurately Tracking Relative Positions on Moving Trackers based on UWB Ranging and Inertial Sensing without Anchors}
}

@InProceedings{Lee_2024_CVPR,
    author    = {Lee, Jiye and Joo, Hanbyul},
    title     = {Mocap Everyone Everywhere: Lightweight Motion Capture With Smartwatches and a Head-Mounted Camera},
    booktitle = {Proceedings of the IEEE/CVF Conference on Computer Vision and Pattern Recognition (CVPR)},
    month     = {June},
    year      = {2024},
    pages     = {1091-1100}
}

@inproceedings{jiang2024egoposer,
  title={EgoPoser: Robust real-time egocentric pose estimation from sparse and intermittent observations everywhere},
  author={Jiang, Jiaxi and Streli, Paul and Meier, Manuel and Holz, Christian},
  booktitle={European Conference on Computer Vision},
  year={2024}
}

@inproceedings{yi2024pnp,
  title={Physical Non-inertial Poser (PNP): Modeling Non-inertial Effects in Sparse-inertial Human Motion Capture},
  author={Yi, Xinyu and Zhou, Yuxiao and Xu, Feng},
  booktitle={SIGGRAPH 2024 Conference Papers},
  year={2024}
}

@InProceedings{Van_Wouwe_2024_CVPR,
    author    = {Van Wouwe, Tom and Lee, Seunghwan and Falisse, Antoine and Delp, Scott and Liu, C. Karen},
    title     = {DiffusionPoser: Real-time Human Motion Reconstruction From Arbitrary Sparse Sensors Using Autoregressive Diffusion},
    booktitle = {Proceedings of the IEEE/CVF Conference on Computer Vision and Pattern Recognition (CVPR)},
    year      = {2024},
    pages     = {2513-2523}
}

@article{liu2024egohdm,
  title={{EgoHDM: An Online Egocentric-Inertial Human Motion Capture, Localization, and Dense Mapping System}},
  author={Liu, Bonan and Yin, Handi and Kaufmann, Manuel and He, Jinhao and Christen, Sammy and Song, Jie and Hui, Pan},
  volume = {43},
  number = {6},
  journal = {ACM Trans. Graph.},  
  year={2024}
}

@INPROCEEDINGS{egohumans,
  author={Khirodkar, Rawal and Bansal, Aayush and Ma, Lingni and Newcombe, Richard and Vo, Minh and Kitani, Kris},
  booktitle={2023 IEEE/CVF International Conference on Computer Vision (ICCV)}, 
  title={{EgoHumans: An Egocentric 3D Multi-Human Benchmark}}, 
  year={2023}
}

@InProceedings{Zuo_2024_CVPR_loose,
    author    = {Zuo, Chengxu and Wang, Yiming and Zhan, Lishuang and Guo, Shihui and Yi, Xinyu and Xu, Feng and Qin, Yipeng},
    title     = {Loose Inertial Poser: Motion Capture with IMU-attached Loose-Wear Jacket},
    booktitle = {Proceedings of the IEEE/CVF Conference on Computer Vision and Pattern Recognition (CVPR)},
    year      = {2024}
}

@inproceedings{jiang2024manikin,
  title={Manikin: biomechanically accurate neural inverse kinematics for human motion estimation},
  author={Jiang, Jiaxi and Streli, Paul and Luo, Xuejing and Gebhardt, Christoph and Holz, Christian},
  booktitle={European Conference on Computer Vision},
  pages={128--146},
  year={2024},
  organization={Springer}
}

@article{ilic2025human,
  title={Human Motion Capture from Loose and Sparse Inertial Sensors with Garment-aware Diffusion Models},
  author={Ilic, Andela and Jiang, Jiaxi and Streli, Paul and Liu, Xintong and Holz, Christian},
  journal={arXiv preprint arXiv:2506.15290},
  year={2025}
}

@inproceedings{shin2024wham,
  title={Wham: Reconstructing world-grounded humans with accurate 3d motion},
  author={Shin, Soyong and Kim, Juyong and Halilaj, Eni and Black, Michael J},
  booktitle={Proceedings of the IEEE/CVF Conference on Computer Vision and Pattern Recognition},
  pages={2070--2080},
  year={2024}
}

@inproceedings{wang2024tram,
  title={TRAM: Global Trajectory and Motion of 3D Humans from in-the-wild Videos},
  author={Wang, Yufu and Wang, Ziyun and Liu, Lingjie and Daniilidis, Kostas},
  booktitle={European Conference on Computer Vision},
  pages={467--487},
  year={2024}
}

@inproceedings{wang2025prompthmr,
  title={PromptHMR: Promptable Human Mesh Recovery},
  author={Wang, Yufu and Sun, Yu and Patel, Priyanka and Daniilidis, Kostas and Black, Michael J and Kocabas, Muhammed},
  booktitle={Proceedings of the Computer Vision and Pattern Recognition Conference},
  pages={1148--1159},
  year={2025}
}

@article{yi2025improving,
  title={Improving Global Motion Estimation in Sparse IMU-based Motion Capture with Physics},
  author={Yi, Xinyu and Pan, Shaohua and Xu, Feng},
  journal={ACM Transactions on Graphics (TOG)},
  volume={44},
  number={4},
  pages={1--16},
  year={2025},
  publisher={ACM New York, NY, USA}
}

@article{starke2024categorical,
  title={Categorical codebook matching for embodied character controllers},
  author={Starke, Sebastian and Starke, Paul and He, Nicky and Komura, Taku and Ye, Yuting},
  journal={ACM Transactions on Graphics (TOG)},
  volume={43},
  number={4},
  pages={1--14},
  year={2024},
  publisher={ACM New York, NY, USA}
}

@inproceedings{guzov2024hmd,
  title={{{HMD}$^2$}: Environment-aware Motion Generation from Single Egocentric Head-Mounted Device},
  author={Guzov, Vladimir and Jiang, Yifeng and Hong, Fangzhou and Pons-Moll, Gerard and Newcombe, Richard and Liu, C Karen and Ye, Yuting and Ma, Lingni},
  booktitle={International Conference on 3D Vision (3DV)},
  year={2025}
}

@inproceedings{ma2024nymeria,
  title={Nymeria: A massive collection of multimodal egocentric daily motion in the wild},
  author={Ma, Lingni and Ye, Yuting and Hong, Fangzhou and Guzov, Vladimir and Jiang, Yifeng and Postyeni, Rowan and Pesqueira, Luis and Gamino, Alexander and Baiyya, Vijay and Kim, Hyo Jin and others},
  booktitle={European Conference on Computer Vision},
  pages={445--465},
  year={2024}
}

@inproceedings{zhang2024rohm,
  title={{RoHM: Robust human motion reconstruction via diffusion}},
  author={Zhang, Siwei and Bhatnagar, Bharat Lal and Xu, Yuanlu and Winkler, Alexander and Kadlecek, Petr and Tang, Siyu and Bogo, Federica},
  booktitle={Proceedings of the IEEE/CVF Conference on Computer Vision and Pattern Recognition},
  pages={14606--14617},
  year={2024}
}

@article{droid,
  title={{DROID-SLAM}: Deep visual slam for monocular, stereo, and {RGB-D} cameras},
  author={Teed, Zachary and Deng, Jia},
  journal={Advances in Neural Information Processing Systems},
  volume={34},
  pages={16558--16569},
  year={2021}
}

@inproceedings{
    daip2024hmdposer,
    title={HMD-Poser: On-Device Real-time Human Motion Tracking from Scalable Sparse Observations},
    author={Dai, Peng and Zhang, Yang and Liu, Tao and Fan, Zhen and Du, Tianyuan and Su, Zhuo and Zheng, Xiaozheng and Li, Zeming},
    booktitle={Proceedings of the IEEE/CVF Conference on Computer Vision and Pattern Recognition},
    year={2024}
}

@article{vitpose,
  title={{VITPose}: Simple vision transformer baselines for human pose estimation},
  author={Xu, Yufei and Zhang, Jing and Zhang, Qiming and Tao, Dacheng},
  journal={Advances in Neural Information Processing Systems},
  volume={35},
  pages={38571--38584},
  year={2022}
}

@inproceedings{yolo,
  title={{YOLOv7}: Trainable bag-of-freebies sets new state-of-the-art for real-time object detectors},
  author={Wang, Chien-Yao and Bochkovskiy, Alexey and Liao, Hong-Yuan Mark},
  booktitle={Proceedings of the IEEE/CVF Conference on Computer Vision and Pattern Recognition},
  pages={7464--7475},
  year={2023}
}

@inproceedings{smplx,
  title={Expressive body capture: {3D} hands, face, and body from a single image},
  author={Pavlakos, Georgios and Choutas, Vasileios and Ghorbani, Nima and Bolkart, Timo and Osman, Ahmed AA and Tzionas, Dimitrios and Black, Michael J},
  booktitle={Proceedings of the IEEE/CVF Conference on Computer Vision and Pattern Recognition},
  pages={10975--10985},
  year={2019}
}

@inproceedings{wang2026embodmocap,
    title = {{EmbodMocap: In-the-Wild 4D Human-Scene Reconstruction for Embodied Agents}},
    booktitle = {Proceedings of the IEEE/CVF Conference on Computer Vision and Pattern Recognition},
    author = {Wang, Wenjia and Pan, Liang and Pi, Huaijin and Lou, Yuke and Ren, Xuqian and Wu, Yifan and Liao, Zhouyingcheng and Yang, Lei and Dabral, Rishabh and Theobalt, Christian and Komura, Taku.},
    year = {2026}
  }

@inproceedings{spin,
  title={Learning to reconstruct {3D} human pose and shape via model-fitting in the loop},
  author={Kolotouros, Nikos and Pavlakos, Georgios and Black, Michael J and Daniilidis, Kostas},
  booktitle={Proceedings of the IEEE/CVF International Conference on Computer Vision},
  pages={2252--2261},
  year={2019}
}

@InProceedings{dsd,
author = {Sun, Yu and Ye, Yun and Liu, Wu and Gao, Wenpeng and Fu, Yili and Mei, Tao},
title = {Human Mesh Recovery From Monocular Images via a Skeleton-Disentangled Representation},
booktitle = {Proceedings of the IEEE/CVF International Conference on Computer Vision},
year = {2019}
}

@inproceedings{refit,
  title={{ReFit}: Recurrent fitting network for {3D} human recovery},
  author={Wang, Yufu and Daniilidis, Kostas},
  booktitle={Proceedings of the IEEE/CVF International Conference on Computer Vision},
  pages={14644--14654},
  year={2023}
}

@inproceedings{hmr,
  title={End-to-end recovery of human shape and pose},
  author={Kanazawa, Angjoo and Black, Michael J and Jacobs, David W and Malik, Jitendra},
  booktitle={Proceedings of the IEEE/CVF Conference on Computer Vision and Pattern Recognition},
  pages={7122--7131},
  year={2018}
}

@article{hmr2,
  title={Reconstructing and Tracking Humans with Transformers},
  author={Goel, Shubham and Pavlakos, Georgios and Rajasegaran, Jathushan and Kanazawa, Angjoo and Malik, Jitendra},
  journal={Proceedings of the IEEE/CVF International Conference on Computer Vision},
  year={2023}
}

@inproceedings{meshgraphform,
  title={Mesh graphormer},
  author={Lin, Kevin and Wang, Lijuan and Liu, Zicheng},
  booktitle={Proceedings of the IEEE/CVF International Conference on Computer Vision},
  pages={12939--12948},
  year={2021}
}

@inproceedings{pare,
  title={{PARE}: Part attention regressor for {3D} human body estimation},
  author={Kocabas, Muhammed and Huang, Chun-Hao P and Hilliges, Otmar and Black, Michael J},
  booktitle={Proceedings of the IEEE/CVF International Conference on Computer Vision},
  pages={11127--11137},
  year={2021}
}

@inproceedings{pymaf,
  title={{PyMAF}: {3D} human pose and shape regression with pyramidal mesh alignment feedback loop},
  author={Zhang, Hongwen and Tian, Yating and Zhou, Xinchi and Ouyang, Wanli and Liu, Yebin and Wang, Limin and Sun, Zhenan},
  booktitle={Proceedings of the IEEE/CVF International Conference on Computer Vision},
  pages={11446--11456},
  year={2021}
}

@inproceedings{cliff,
  title={{CLIFF}: Carrying location information in full frames into human pose and shape estimation},
  author={Li, Zhihao and Liu, Jianzhuang and Zhang, Zhensong and Xu, Songcen and Yan, Youliang},
  booktitle={European Conference on Computer Vision},
  pages={590--606},
  year={2022}
}

@inproceedings{hmmr,
  title={Learning {3D} human dynamics from video},
  author={Kanazawa, Angjoo and Zhang, Jason Y and Felsen, Panna and Malik, Jitendra},
  booktitle={Proceedings of the IEEE/CVF Conference on Computer Vision and Pattern Recognition},
  pages={5614--5623},
  year={2019}
}

@inproceedings{vibe,
  title={{VIBE}: Video inference for human body pose and shape estimation},
  author={Kocabas, Muhammed and Athanasiou, Nikos and Black, Michael J},
  booktitle={Proceedings of the IEEE/CVF Conference on Computer Vision and Pattern Recognition},
  pages={5253--5263},
  year={2020}
}

@inproceedings{tcmr,
  title={Beyond static features for temporally consistent {3D} human pose and shape from a video},
  author={Choi, Hongsuk and Moon, Gyeongsik and Chang, Ju Yong and Lee, Kyoung Mu},
  booktitle={Proceedings of the IEEE/CVF Conference on Computer Vision and Pattern Recognition},
  pages={1964--1973},
  year={2021}
}

@inproceedings{I2l,
  title={{I2L-MeshNet}: Image-to-lixel prediction network for accurate 3d human pose and mesh estimation from a single {RGB} image},
  author={Moon, Gyeongsik and Lee, Kyoung Mu},
  booktitle={European Conference on Computer Vision},
  pages={752--768},
  year={2020}
}

@inproceedings{pose2mesh,
  title={{Pose2Mesh}: Graph convolutional network for {3D} human pose and mesh recovery from a {2D} human pose},
  author={Choi, Hongsuk and Moon, Gyeongsik and Lee, Kyoung Mu},
  booktitle={European Conference on Computer Vision},
  pages={769--787},
  year={2020},
  organization={Springer}
}

@article{multihmr,
  title={{Multi-HMR}: Multi-Person Whole-Body Human Mesh Recovery in a Single Shot},
  author={Baradel, Fabien and Armando, Matthieu and Galaaoui, Salma and Br{\'e}gier, Romain and Weinzaepfel, Philippe and Rogez, Gr{\'e}gory and Lucas, Thomas},
  journal={European Conference on Computer Vision},
  year={2024}
}

@inproceedings{slahmr,
  title={Decoupling human and camera motion from videos in the wild},
  author={Ye, Vickie and Pavlakos, Georgios and Malik, Jitendra and Kanazawa, Angjoo},
  booktitle={Proceedings of the IEEE/CVF Conference on Computer Vision and Pattern Recognition},
  pages={21222--21232},
  year={2023}
}

@inproceedings{liu20214d,
  title={4d human body capture from egocentric video via 3d scene grounding},
  author={Liu, Miao and Yang, Dexin and Zhang, Yan and Cui, Zhaopeng and Rehg, James M and Tang, Siyu},
  booktitle={2021 international conference on 3D vision (3DV)},
  pages={930--939},
  year={2021},
  organization={IEEE}
}

@inproceedings{pace,
  title={{PACE}: Human and Camera Motion Estimation from in-the-wild Videos},
  author={Kocabas, Muhammed and Yuan, Ye and Molchanov, Pavlo and Guo, Yunrong and Black, Michael J and Hilliges, Otmar and Kautz, Jan and Iqbal, Umar},
  booktitle={International Conference on 3D Vision},
  pages={397--408},
  year={2024},
}

@inproceedings{bodyslam,
  title={{BodySLAM}: joint camera localisation, mapping, and human motion tracking},
  author={Henning, Dorian F and Laidlow, Tristan and Leutenegger, Stefan},
  booktitle={European Conference on Computer Vision},
  pages={656--673},
  year={2022}
}

@inproceedings{bodyslam++,
  title={{BodySLAM++}: Fast and Tightly-Coupled Visual-Inertial Camera and Human Motion Tracking},
  author={Henning, Dorian F and Choi, Christopher and Schaefer, Simon and Leutenegger, Stefan},
  booktitle={IEEE/RSJ International Conference on Intelligent Robots and Systems},
  pages={3781--3788},
  year={2023},
  organization={IEEE}
}

@inproceedings{glamr,
  title={{GLAMR}: Global occlusion-aware human mesh recovery with dynamic cameras},
  author={Yuan, Ye and Iqbal, Umar and Molchanov, Pavlo and Kitani, Kris and Kautz, Jan},
  booktitle={Proceedings of the IEEE/CVF Conference on Computer Vision and Pattern Recognition},
  pages={11038--11049},
  year={2022}
}

@inproceedings{multiphys,
  title={{MultiPhys}: Multi-Person Physics-aware {3D} Motion Estimation},
  author={Ugrinovic, Nicolas and Pan, Boxiao and Pavlakos, Georgios and Paschalidou, Despoina and Shen, Bokui and Sanchez-Riera, Jordi and Moreno-Noguer, Francesc and Guibas, Leonidas},
  booktitle={Proceedings of the IEEE/CVF Conference on Computer Vision and Pattern Recognition},
  pages={2331--2340},
  year={2024}
}

@inproceedings{castillo2023bodiffusion,
  title={Bodiffusion: Diffusing sparse observations for full-body human motion synthesis},
  author={Castillo, Angela and Escobar, Maria and Jeanneret, Guillaume and Pumarola, Albert and Arbel{\'a}ez, Pablo and Thabet, Ali and Sanakoyeu, Artsiom},
  booktitle={Proceedings of the IEEE/CVF International Conference on Computer Vision},
  pages={4221--4231},
  year={2023}
}

@inproceedings{genmo2025,
  title={{GENMO: Generative Models for Human Motion Synthesis}},
  author={Li, Jiefeng and Cao, Jinkun and Zhang, Haotian and Rempe, Davis and Kautz, Jan and Iqbal, Umar and Yuan, Ye},
  booktitle={Proceedings of the IEEE/CVF International Conference on Computer Vision},
  year={2025}
}

@inproceedings{harmony4d,
    author = {Khirodkar, Rawal and Song, Jyun-Ting and Cao, Jinkun and Luo, Zhengyi and Kitani, Kris},
    title = {{Harmony4D: a video dataset for in-the-wild close human interactions}},
    year = {2024},    
    booktitle = {Proceedings of the 38th International Conference on Neural Information Processing Systems},    
}

@inproceedings{egolm,
    title={{EgoLM: Multi-Modal Language Model of Egocentric Motions}},
    author={Fangzhou Hong and Vladimir Guzov and Hyo Jin Kim and Yuting Ye and Richard Newcombe and Ziwei Liu and Lingni Ma},
    booktitle={Proceedings of the IEEE/CVF Conference on Computer Vision and Pattern Recognition},
    year={2025}
}

@InProceedings{Zhao2024globalcamhuman,
    author    = {Zhao, Yizhou and Wang, Tuanfeng Yang and Raj, Bhiksha and Xu, Min and Yang, Jimei and Huang, Chun-Hao Paul},
    title     = {Synergistic Global-space Camera and Human Reconstruction from Videos},
    booktitle = {Proceedings of the IEEE/CVF Conference on Computer Vision and Pattern Recognition},
    year      = {2024},
    pages     = {1216-1226}
}

@inproceedings{shen2024gvhmr,
  title={World-Grounded Human Motion Recovery via Gravity-View Coordinates},
  author={Shen, Zehong and Pi, Huaijin and Xia, Yan and Cen, Zhi and Peng, Sida and Hu, Zechen and Bao, Hujun and Hu, Ruizhen and Zhou, Xiaowei},
  booktitle={SIGGRAPH Asia},
  year={2024}
}

@inproceedings{SPEC:ICCV:2021,
  title = {{SPEC}: Seeing People in the Wild with an Estimated Camera},
  author = {Kocabas, Muhammed and Huang, Chun-Hao P. and Tesch, Joachim and Müller, Lea and Hilliges, Otmar and Black, Michael J.},
  booktitle = {Proceedings of the IEEE/CVF International Conference on Computer Vision},
  pages = {11035--11045},
  year = {2021}
}

@article{teed2024dpvo,
  title={Deep patch visual odometry},
  author={Teed, Zachary and Lipson, Lahav and Deng, Jia},
  journal={Advances in Neural Information Processing Systems},
  volume={36},
  year={2024}
}

@InProceedings{Luo_2020_ACCV,
    author    = {Luo, Zhengyi and Golestaneh, S. Alireza and Kitani, Kris M.},
    title     = {3D Human Motion Estimation via Motion Compression and Refinement},
    booktitle = {Proceedings of the Asian Conference on Computer Vision},
    year      = {2020}
}

@inproceedings{yi2025estimating,
  title={Estimating body and hand motion in an ego-sensed world},
  author={Yi, Brent and Ye, Vickie and Zheng, Maya and Li, Yunqi and M{\"u}ller, Lea and Pavlakos, Georgios and Ma, Yi and Malik, Jitendra and Kanazawa, Angjoo},
  booktitle={Proceedings of the Computer Vision and Pattern Recognition Conference},
  pages={7072--7084},
  year={2025}
}

@inproceedings{barquero2025sparse,
  title={From Sparse Signal to Smooth Motion: Real-Time Motion Generation with Rolling Prediction Models},
  author={Barquero, German and Bertsch, Nadine and Marramreddy, Manojkumar and Chac{\'o}n, Carlos and Arcadu, Filippo and Rigual, Ferran and He, Nicky Sijia and Palmero, Cristina and Escalera, Sergio and Ye, Yuting and others},
  booktitle={Proceedings of the Computer Vision and Pattern Recognition Conference},
  pages={1850--1860},
  year={2025}
}

@article{engel2023project,
  title={Project Aria: A new tool for egocentric multi-modal AI research},
  author={Engel, Jakob and Somasundaram, Kiran and Goesele, Michael and Sun, Albert and Gamino, Alexander and Turner, Andrew and Talattof, Arjang and Yuan, Arnie and Souti, Bilal and Meredith, Brighid and others},
  journal={arXiv preprint arXiv:2308.13561},
  year={2023}
}

@article{chen2024motion,
  title={Motion capture from inertial and vision sensors},
  author={Chen, Xiaodong and Liu, Wu and Bao, Qian and Liu, Xinchen and Yang, Quanwei and Dai, Ruoli and Mei, Tao},
  journal={arXiv preprint arXiv:2407.16341},
  year={2024}
}

@inproceedings{zhang2024mmvp,
  title={Mmvp: A multimodal mocap dataset with vision and pressure sensors},
  author={Zhang, He and Ren, Shenghao and Yuan, Haolei and Zhao, Jianhui and Li, Fan and Sun, Shuangpeng and Liang, Zhenghao and Yu, Tao and Shen, Qiu and Cao, Xun},
  booktitle={Proceedings of the IEEE/CVF Conference on Computer Vision and Pattern Recognition},
  pages={21842--21852},
  year={2024}
}

@inproceedings{yan2024reli11d,
  title={Reli11d: A comprehensive multimodal human motion dataset and method},
  author={Yan, Ming and Zhang, Yan and Cai, Shuqiang and Fan, Shuqi and Lin, Xincheng and Dai, Yudi and Shen, Siqi and Wen, Chenglu and Xu, Lan and Ma, Yuexin and others},
  booktitle={Proceedings of the IEEE/CVF Conference on Computer Vision and Pattern Recognition},
  pages={2250--2262},
  year={2024}
}

@inproceedings{liu2025umotion,
  title={UMotion: Uncertainty-driven Human Motion Estimation from Inertial and Ultra-wideband Units},
  author={Liu, Huakun and Ota, Hiroki and Wei, Xin and Hirao, Yutaro and Perusquia-Hernandez, Monica and Uchiyama, Hideaki and Kiyokawa, Kiyoshi},
  booktitle={Proceedings of the Computer Vision and Pattern Recognition Conference},
  pages={7085--7094},
  year={2025}
}

@article{simeoni2025dinov3,
  title={Dinov3},
  author={Sim{\'e}oni, Oriane and Vo, Huy V and Seitzer, Maximilian and Baldassarre, Federico and Oquab, Maxime and Jose, Cijo and Khalidov, Vasil and Szafraniec, Marc and Yi, Seungeun and Ramamonjisoa, Micha{\"e}l and others},
  journal={arXiv preprint arXiv:2508.10104},
  year={2025}
}

@misc{apple_vision_pro,
  title = {Apple {Vision Pro}},
  year = {2024},
  howpublished = {\url{https://www.apple.com/apple-vision-pro/}},
  note = {Accessed: 2026-02-28}
}

@misc{hololens,
  title = {Microsoft {HoloLens} 2},
  year = {2019},
  howpublished = {\url{https://www.microsoft.com/hololens}},
  note = {Accessed: 2026-02-28}
}

@misc{quest,
  title = {Meta {Quest} 3},
  year = {2023},
  howpublished = {\url{https://www.meta.com/quest/quest-3/}},
  note = {Accessed: 2026-02-28}
}

@inproceedings{zhan2022ray3d,
  title={Ray3d: ray-based 3d human pose estimation for monocular absolute 3d localization},
  author={Zhan, Yu and Li, Fenghai and Weng, Renliang and Choi, Wongun},
  booktitle={Proceedings of the IEEE/CVF conference on computer vision and pattern recognition},
  pages={13116--13125},
  year={2022}
}

@article{pujades2019virtual,
  title={The Virtual Caliper: Rapid Creation of Metrically Accurate Avatars from 3D Measurements},
  author={Pujades, Sergi and Mohler, Betty and Thaler, Anne and Tesch, Joachim and Mahmood, Naureen and Hesse, Nikolas and B{\"u}lthoff, Heinrich H and Black, Michael J},
  journal={IEEE transactions on visualization and computer graphics},
  volume={25},
  number={5},
  pages={1887--1897},
  year={2019},
  publisher={IEEE},
   
}

@inproceedings{xue2025group,
  title={Group Inertial Poser: Multi-Person Pose and Global Translation from Sparse Inertial Sensors and Ultra-Wideband Ranging},
  author={Xue, Ying and Jiang, Jiaxi and Armani, Rayan and Hollidt, Dominik and Liao, Yi-Chi and Holz, Christian},
  booktitle={Proceedings of the IEEE/CVF International Conference on Computer Vision},
  pages={24910--24921},
  year={2025}
}
              }

              \end{document}